\documentclass[letterpaper, 10 pt, conference]{ieeeconf}  
\pdfoutput=1

\IEEEoverridecommandlockouts                              

\usepackage{graphicx} 
\usepackage{amsmath} 
\usepackage{amssymb}  
\usepackage{bm}
\usepackage{subfigure}
\usepackage{color}
\definecolor{mygray}{gray}{0.6}
\usepackage{cite}
\usepackage[linesnumbered,ruled]{algorithm2e}
\usepackage{hyperref}
\usepackage{float}
\usepackage{booktabs}
\usepackage{multirow}
\usepackage{url}
\usepackage{comment}
\usepackage[utf8]{inputenc}
\hypersetup{
	colorlinks=true,
	linkcolor=blue,
	filecolor=magenta,      
	urlcolor=cyan,
}
\usepackage{pifont} 
\newcommand{\xmark}{\text{\ding{55}}}
\newcommand{\cmark}{\ding{51}}%

\def\figureautorefname~#1\null{Fig.~~#1\null}
\def\equationautorefname~#1\null{Eq.~~(#1)\null}

\title{\LARGE \bf 
MULLS: Versatile LiDAR SLAM via Multi-metric Linear Least Square
}

\author{Yue Pan$^{1}$, Pengchuan Xiao$^{2}$, Yujie He$^{3}$, Zhenlei Shao$^{2}$ and  Zesong Li$^{2}$ 
\thanks{$^{1}$Y.Pan is with Department of Civil, Environmental and Geomatic Engineering,
        ETH Zurich, Switzerland
        {\tt\small yuepan@ethz.ch}}%
\thanks{$^{2}$P.Xiao, Z.Shao and Z.Li are with Hesai Technology Co., Ltd., Shanghai, China {\tt\small
\{xiaopengchuan\_intern, shaozhenlei, lizesong\}@hesaitech.com}}
\thanks{$^{3}$Y.He is with the School of Engineering, EPFL, Lausanne, Switzerland
        {\tt\small yujie.he@epfl.ch}}%
}

\begin{document}
	
	\maketitle
	\thispagestyle{empty}
	\pagestyle{empty}

	\begin{abstract}
	The rapid development of autonomous driving and mobile mapping calls for off-the-shelf LiDAR SLAM solutions that are adaptive to LiDARs of different specifications on various complex scenarios. To this end, we propose MULLS, an efficient, low-drift, and versatile 3D LiDAR SLAM system. For the front-end, roughly classified feature points (ground, facade, pillar, beam, etc.) are extracted from each frame using dual-threshold ground filtering and principal components analysis. Then the registration between the current frame and the local submap is accomplished efficiently by the proposed multi-metric linear least square iterative closest point algorithm. Point-to-point (plane, line) error metrics within each point class are jointly optimized with a linear approximation to estimate the ego-motion.  Static feature points of the registered frame are appended into the local map to keep it updated. For the back-end, hierarchical pose graph optimization is conducted among regularly stored history submaps to reduce the drift resulting from dead reckoning. Extensive experiments are carried out on three datasets with more than 100,000 frames collected by seven types of LiDAR on various outdoor and indoor scenarios. On the KITTI benchmark, MULLS ranks among the top LiDAR-only SLAM systems with real-time performance.

	\end{abstract}
	
	
	
	\section{Introduction}\label{sec:intro}
Simultaneous localization and mapping (SLAM) plays a key role in robotics tasks, including robot navigation\cite{temeltas2008slam}, field surveying\cite{ebadi2020lamp} and high-definition map production\cite{ma2019exploiting} for autonomous driving. Compared with vision-based \cite{mur2015orb,forster2014svo,qin2018vins} and RGB-D-based\cite{izadi2011kinectfusion,whelan2015real,labbe2019rtab} SLAM systems, LiDAR SLAM systems are more reliable under various lighting conditions and are more suitable for tasks demanding for dense 3D map. 


The past decade has witnessed enormous development in LiDAR odometry (LO)\cite{zhang2014loam,deschaud2018imls,graeter2018limo,neuhaus2018mc2slam,shan2018lego,behley2018efficient,chen2019suma++,kovalenko2019sensor,li2019net,ye2019tightly,zuo2019lic,lin2020livox,chen2020sloamforest,aloam,liosam2020shan,garcia2020fail,zhou2020s4,chen2020psf,zheng2020lodonet}, LiDAR-based loop closure detection\cite{he2016m2dp,dube2020segmap,kim2018scan,wang2020isc,jiang2020lipmatch,liang2020novel,chen2020overlapnet,zaganidis2019semantically} and pose optimization\cite{kaess2008isam,grisetti2010hierarchical,ni2010multilevelsubmap,kummerle2011g,grisetti2012robustoptimizationofgraph, Blanco-Claraco-RSS-19}. Though state-of-the-art LiDAR-based approaches performs well in structured urban or indoor scenes with the localization accuracy better than most of the vision-based methods, the lack of versatility hinders them from being widely used in the industry. Most of methods are based on sensor dependent point cloud representations such as, ring\cite{zhang2014loam,zuo2019lic} and range image\cite{shan2018lego,behley2018efficient,chen2019suma++,kovalenko2019sensor,li2019net}, which requires detailed LiDAR model parameters such as, scan-line distribution and the field of view. It's non-trivial to directly adapt them on recently developed mechanical LiDARs with unique configuration of beams and solid-state LiDARs with limited field of view\cite{lin2020livox}. Besides, some SLAM systems target on specific application scenarios such as, urban road\cite{li2020urbanslamisprs}, highway\cite{zhao2019highway}, tunnel\cite{palieri2020locus} and forest\cite{chen2020sloamforest} using the ad-hoc framework. They tend to get stuck in degenerated corner-cases or a rapid switch of the scene.




	\begin{figure}[t]
		\centering
	\setlength{\abovecaptionskip}{-16pt}
		\includegraphics[width=0.48\textwidth]{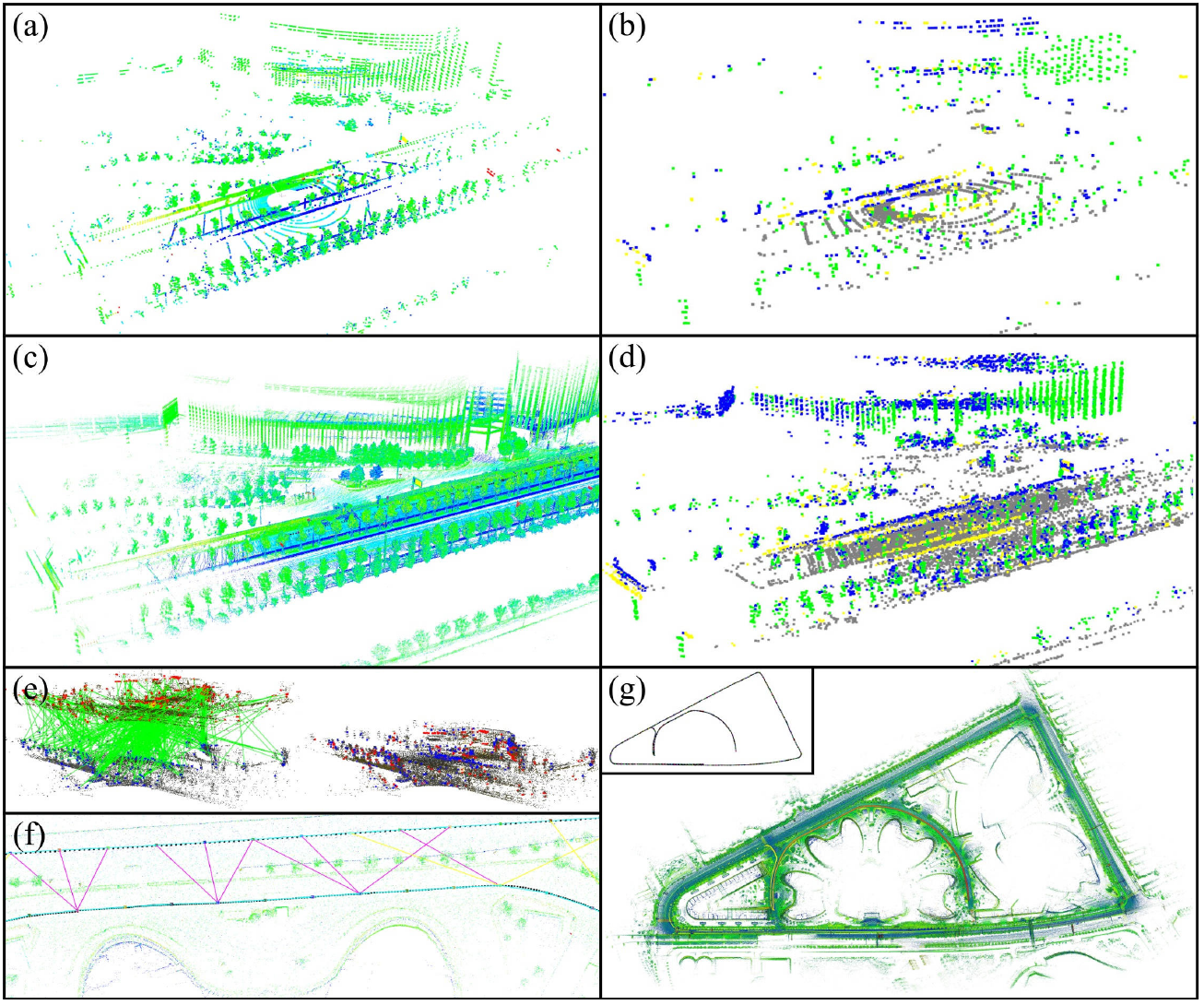}
		\caption{Overview of the proposed MULLS-SLAM system: (a) the point cloud from a single scan of a multi-beam spinning LiDAR, (b) various types of feature points (\textcolor{mygray}{ground}, \textcolor{blue}{facade}, \textcolor{green}{pillar}, \textcolor{yellow}{beam}) extracted from the scan, (c) registered point cloud of the local map, (d) feature points of the local map, (e) global registration between revisited submaps by TEASER\cite{Yang20tro-teaser}, (f) loop closure edges among submaps, (g) generated consistent map and trajectory.}
		\label{fig:overview}
	\vspace{-16pt}
	\end{figure}


In this work, we proposed a LiDAR-only SLAM system (as shown in Fig.~\ref{fig:overview}) to cope with the aforementioned challenges. The continuous inputs of the system are sets of 3D coordinates (optionally with point-wise intensity and timestamp) without the conversion to rings or range images. Therefore the system is independent of the LiDAR's specification and without the loss of 3D structure information. Geometric feature points with plentiful categories such as ground, facade, pillar, beam, and roof are extracted from each frame, giving rise to better adaptability to the surroundings. 
The ego-motion is estimated efficiently by the proposed multi-metric linear least square (MULLS) iterative closest points (ICP) algorithm based on the classified feature points. 
Furthermore, a submap-based pose graph optimization (PGO) is employed as the back-end with loop closure constraints constructed via map-to-map global registration.
	
The main contributions of this work are listed as follows:
\begin{itemize}
\item A scan-line independent LiDAR-only SLAM solution named MULLS\footnote{\url{https://github.com/YuePanEdward/MULLS}}, with low drift and real-time performance on various scenarios. Currently, MULLS ranks top 10 on the competitive KITTI benchmark\footnote{\url{http://www.cvlibs.net/datasets/kitti/eval\_odometry.php}}.
\item An efficient point cloud local registration algorithm named MULLS-ICP that realizes the linear least square optimization of point-to-point (plane,  line) error metrics jointly in roughly classified geometric feature points. 
\end{itemize}
	
	

	\section{Related works}
	\label{sec:related}
   As the vital operation of LiDAR SLAM, point cloud registration can be categorized into local and global registration. Local registration requires good transformation initial guess to finely align two overlapping scans without stuck in the local minima while global registration can coarsely align them regardless of their relative pose.   
	
	Local registration based on ICP\cite{besl1992method} has been widely used in LO to estimate the relatively small scan-to-scan(map) transformation. 
	Classic ICP keeps alternately solving for dense point-to-point closest correspondences and rigid transformation. To guarantee LO's real-time performance (mostly at 10Hz) on real-life regularized scenarios, faster and more robust variants of ICP have been proposed in recent years focusing on the sampled points for matching and the error metrics\cite{rusinkiewicz2001pointplaneicp,segal2009generalized,censi2008icp,yokozuka2021litamin2} for optimization. As a pioneering work of sparse point-based LO, LOAM\cite{zhang2014loam} selects the edge and planar points by sorting the curvature on each scan-line. The squared sum of point-to-line(plane) distance is minimized by non-linear optimization to estimate the transformation. As a follow-up, LeGO-LOAM\cite{shan2018lego} conducts ground segmentation and isotropic edge points extraction from the projected range image. Then a two-step non-linear optimization is executed on ground and edge correspondences to solve two sets of transformation coefficients successively. Unlike LOAM, SuMa\cite{behley2018efficient} is based on the dense projective normal ICP between the range image of the current scan and the surfel map. SuMa++\cite{chen2019suma++} further improved SuMa by realizing dynamic filtering and multi-class projective matching with the assistance of semantic masks. 
    These methods share the following limits. First, they require the LiDAR's model and lose 3D information due to the operation based on range image or scan-line. \cite{deschaud2018imls}\cite{zhou2020s4} operate based on raw point cloud but lose efficiency. Second, ICP with different error metrics is solved by less efficient non-linear optimization. Though efficient linear least square optimization for point-to-plane metric has been applied on LO\cite{low2004linear,pomerleau2015review,kuhner2020large}, the joint linear solution to point-to-point (plane, line) metrics has not been proposed. These limits are solved in our work.
	
	Global registration is often required for the constraint construction in LiDAR SLAM's back-end, especially when the odometry drifts too much to close the loop by ICP.
	One common solution is to apply correlation operations on global features\cite{kim2018scan,wang2020isc,jiang2020lipmatch,liang2020novel,chen2020overlapnet} encoded from a pair of scans to estimate relative azimuth. 
	Other solutions solve the 6DOF transformation based on local feature matching and verification among the keypoints\cite{rusu2009fpfh,huang2020predator} or segments\cite{dube2020segmap}. Our system follows these solutions by adopting the recently proposed TEASER\cite{Yang20tro-teaser} algorithm. By applying truncated least square estimation with semi-definite relaxation\cite{yang2020graduated}, TEASER is more efficient and robust than methods based on RANSAC\cite{rusu2009fpfh,mellado2014super} and branch \& bound (BnB)\cite{yang2015go,cai2019practical}.
	
	
	\section{Methodology}
	\label{sec:method}
	
	\subsection{Motion compensation}
	\label{sec:undistort}
	Given the point-wise timestamp of a frame, the time ratio for a point $\mathbf{p}_{i}$ with timestamp $\tau_i$ is $s_{i}=\frac{\tau_e-\tau_i}{\tau_e-\tau _b}$, where $\tau_b$, $\tau_e$ are the timestamp at the start and the end of the frame. When IMU is not available, the transformation of $\mathbf{p}_{i}$ to $\mathbf{p}_{i}^{e}$ can be computed under the assumption of uniform motion: 
\begin{equation}
\footnotesize
\mathbf{p}_{i}^{\left( e \right)}=\mathbf{R}_{e,i}\mathbf{p}_{i}+\mathbf{t}_{e,i}\approx \text{slerp}\left( \mathbf{R}_{e,b},s \right) \mathbf{p}_{i}+s\mathbf{t}_{e,b}\label{eq:motion_compensation}
\ ,
\end{equation}
\noindent where $\text{slerp}$ represents the spherical linear interpolation, $\mathbf{R}_{e,b}$ and $\mathbf{t}_{e,b}$ stands for the rotation and translation estimation throughout the frame. The undistorted point cloud is achieved by packing all the points $\{\mathbf{p}_{i}^{e}\}$ at the end of the frame.

\subsection{Geometric feature points extraction and encoding}
\label{sec:geoextract}

\begin{figure}[]
	\centering
	\setlength{\abovecaptionskip}{-12pt}
	\includegraphics[width=0.48\textwidth]{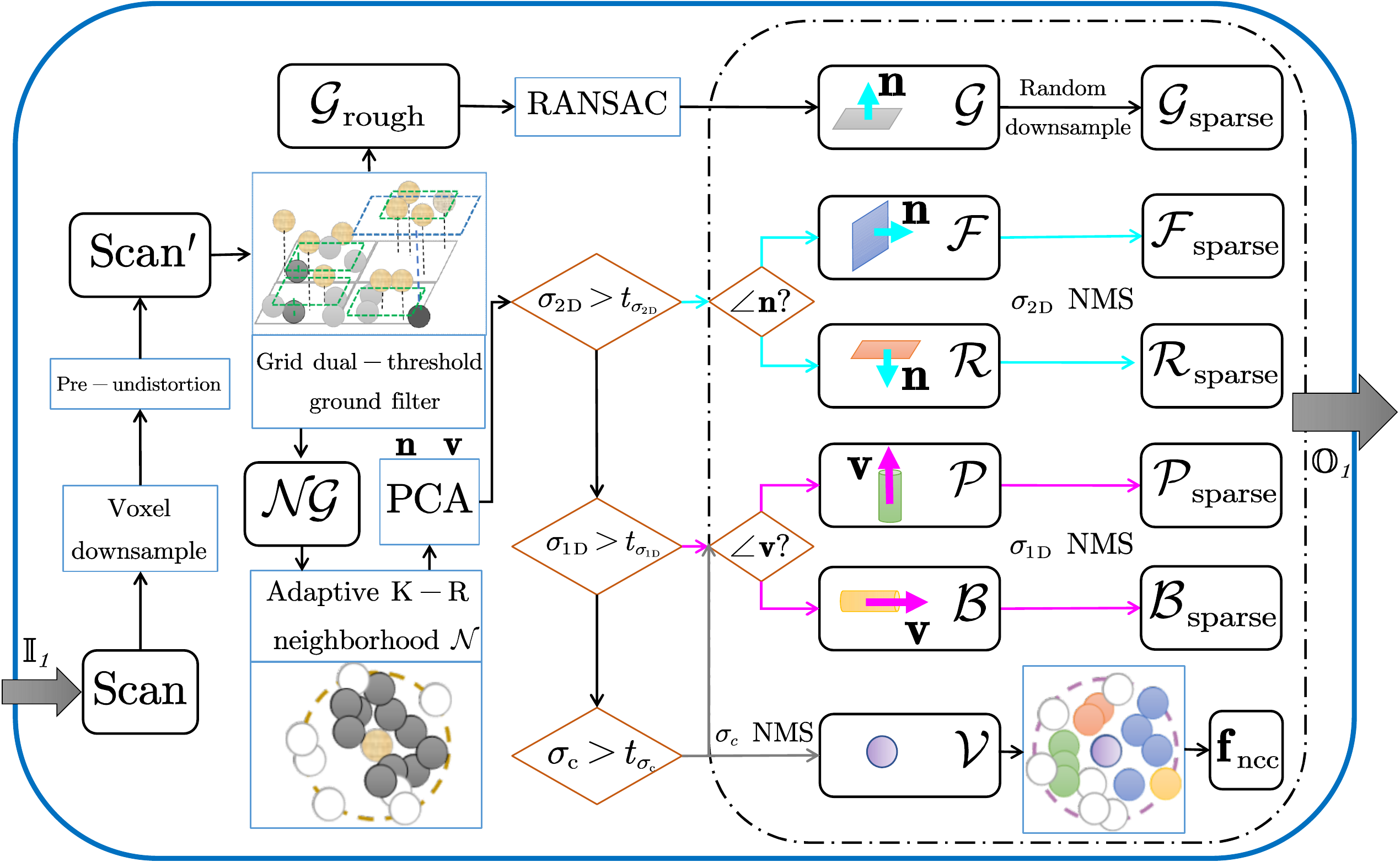}
	\caption{Pipeline of geometric feature points extraction and encoding}
	\label{fig:geofeature}
	\vspace{-12pt}
\end{figure}

The workflow of this section is summarized in Fig.~\ref{fig:geofeature}. The input is the raw point cloud of each frame, and the outputs are six types of feature points with primary and normal vectors.  
\subsubsection{\textbf{\small{Dual-threshold ground filtering}}}
The point cloud after optional preprocessing is projected to a reference plane, which is horizontal or fitted from last frame's ground points. For non-horizontal LiDAR, the initial orientation needs to be known. The reference plane is then divided into equal-size 2D grids. The minimum point height in each grid $g_{i}$ and in its $\small 3\times3$ neighboring grids, denoted as $\footnotesize{ h^{(i)}_{\text{min}}}$ and $\footnotesize{ h^{(i)}_{\text{neimin}}}$, are recorded respectively. With two thresholds $\delta h_1$, $\delta h_2$, each point $p_{k}$ in grid $g_{i}$ is classified as a roughly determined ground point $\mathcal{G}_{\text{rough}}$ or a nonground point $\mathcal{NG}$ by:
	\begin{equation}
	\footnotesize
	 p_{k}^{\left( i \right)}\in \hspace{-2pt}
	 \begin{cases}
	    \begin{aligned}
	    &\mathcal{NG},&&\hspace{-8pt}\text{if } h_k-h_{\text{min}}^{\left( i \right)}>\delta h_1 \text{ or } h_{\text{min}}^{\left( i \right)}-h_{\text{neimin}}^{\left( i \right)}>\delta h_2\\
	    &\mathcal{G}_{\text{rough}},&&\hspace{-8pt}\text{otherwise}\\
	    \end{aligned}
    \end{cases}
    \ .
    \label{eq:ground&nonground}
	\end{equation}
	
	To refine the ground points, RANSAC is employed in each grid to fit the grid-wise ground plane. Inliers are kept as ground points $\mathcal{G}$ and their normal vectors $\mathbf{n}$ are defined as the surface normal of the grid-wise best-fit planes.
    
    \subsubsection{\textbf{\small{Nonground points classification based on PCA}}} Nonground points are downsampled to a fixed number and then fed into the principal components analysis (PCA) module in parallel. The point-wise K-R neighborhood $\mathcal{N}$ is defined as the nearest $K$ points within a sphere of radius $R$. The covariance matrix $\mathbf{C}$ for $\mathcal{N}$ is calculated as $\footnotesize
    \mathbf{C}=\frac{1}{|\mathcal{N}|}\sum_{i\in \mathcal{N}}{\left( \mathbf{p}_i-\mathbf{\bar{p}} \right)}\left( \mathbf{p}_i-\mathbf{\bar{p}} \right) ^\mathsf{T}$, where $\bar{\mathbf{p}}$ is the center of gravity for $\mathcal{N}$. Next, eigenvalues $\footnotesize{\lambda _1>\lambda _2>\lambda_3}$ and the corresponding eigenvectors $\boldsymbol{v},\boldsymbol{m},\boldsymbol{n}$ are determined by the eigenvalue decomposition of $\mathbf{C}$, where $\boldsymbol{v}, \boldsymbol{n}$ are the primary and the normal vector of $\mathcal{N}$. Then the local linearity $\small \sigma _{\text{1D}}$, planarity $\small \sigma _{\text{2D}}$, and curvature $\small \sigma _{\text{c}}$ \cite{hackel2016fast} are defined as $\footnotesize
\sigma _{\text{1D}}=\frac{\lambda _1-\lambda _2}{\lambda _1}, \sigma _{\text{2D}}=\frac{\lambda _2-\lambda _3}{\lambda _1}, \sigma _{\text{c}}=\frac{\lambda _3}{\lambda _1+\lambda _2+\lambda _3}$.

	According to the magnitude of local feature $\sigma _{\text{1D}},\sigma _{\text{2D}},\sigma _{\text{c}}$ and the direction of $\boldsymbol{v}, \boldsymbol{n}$, five categories of feature points can be extracted, namely facade $\mathcal{F}$, roof $\mathcal{R}$, pillar $\mathcal{P}$, beam $\mathcal{B}$, and vertex $\mathcal{V}$. To refine the feature points, non-maximum suppression (NMS) based on $\sigma _{\text{1D}},\sigma _{\text{2D}},\sigma _{\text{c}}$ are applied on linear points ($\mathcal{P}$, $\mathcal{B}$), planar points ($\mathcal{F}$, $\mathcal{R}$) and vertices $\mathcal{V}$ respectively, followed by an isotropic downsampling. Together with $\mathcal{G}$, the roughly classified feature points are packed for registration.
	 
	 \subsubsection{\textbf{\small{Neighborhood category context encoding}}}
	 Based on the extracted feature points, the neighborhood category context (NCC) is proposed to describe each vertex keypoint with almost no other computations roughly. As shown in (\ref{eq:ncc}), the proportion of feature points with different categories in the neighborhood, the normalized intensity and height above ground are encoded. NCC is later used as the local feature for global registration in the SLAM system's back-end (\ref{sec:backend}):
	\begin{equation}
	\footnotesize
\mathbf{f}^{\text{(ncc)}}_{\mathbf{i}}=\left[ \begin{matrix}
	\frac{\left| \mathcal{F}_{\mathcal{N}} \right|}{\left| \mathcal{N} \right|}&		\frac{\left| \mathcal{P}_{\mathcal{N}} \right|}{\left| \mathcal{N} \right|}&		\frac{\left| \mathcal{B}_{\mathcal{N}} \right|}{\left| \mathcal{N} \right|}&		\frac{\left| \mathcal{R}_{\mathcal{N}} \right|}{\left| \mathcal{N} \right|}&	\frac{\bar{I}_{\mathcal{N}}}{I_{\max}}&\frac{h_{\mathcal{G}}}{h_{\max}}  \\
\end{matrix} \right] _{i}^\mathsf{T}
\label{eq:ncc}
\ .
\end{equation}

	\subsection{Multi-metric linear least square ICP}
	\label{sec:mullsicp}
\begin{figure}[]
		\centering
	\setlength{\abovecaptionskip}{-2pt}
		\includegraphics[width=0.47\textwidth]{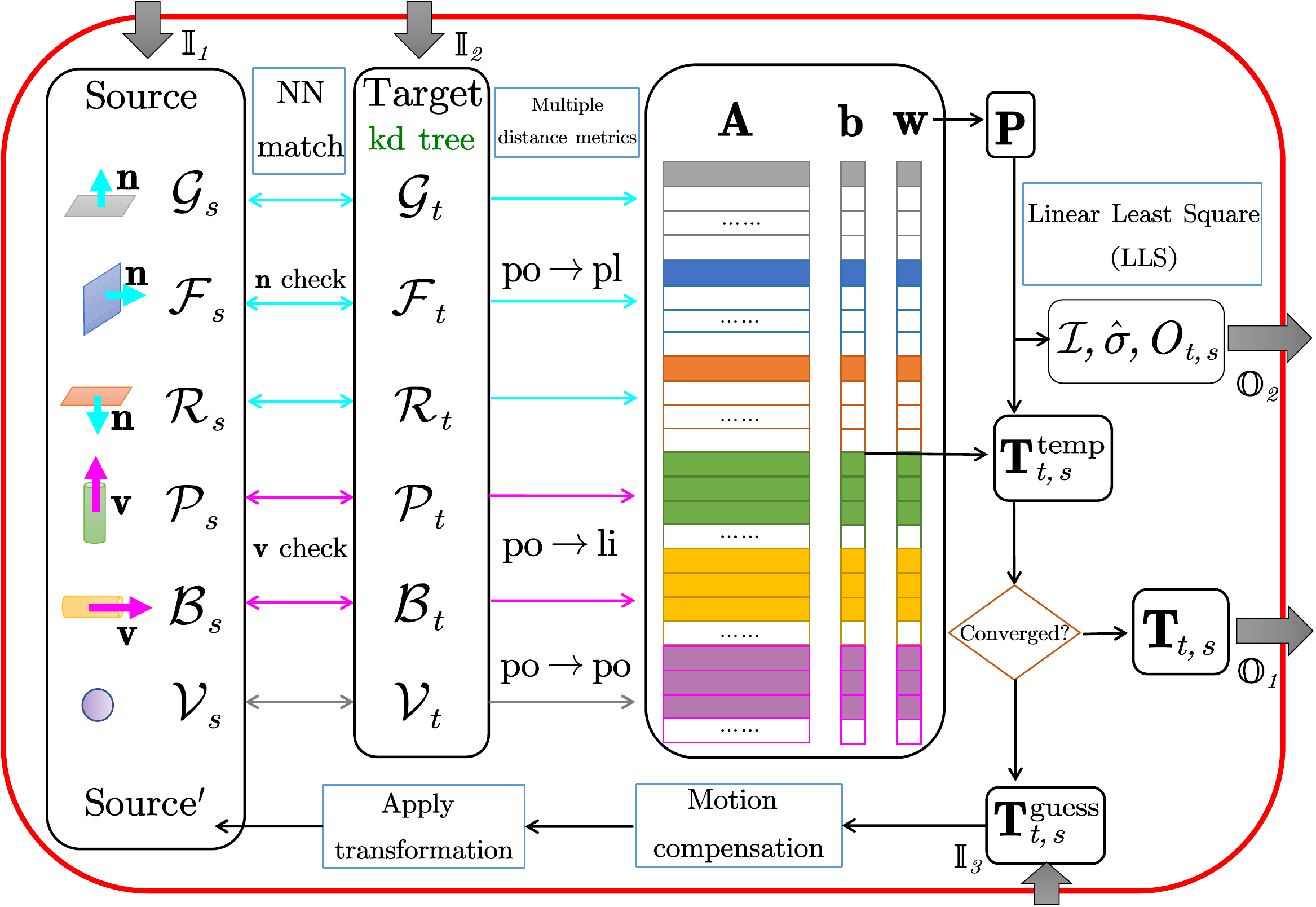}
		\caption{Pipeline of the multi-metric linear least square ICP}
		\label{fig:mulls-icp}
	\vspace{-18pt}
	\end{figure}

    This section's pipeline is summarized in Fig.~\ref{fig:mulls-icp}. The inputs are the source and the target point cloud composed of multi-class feature points extracted in \ref{sec:geoextract}, as well as the initial guess of the transformation from the source point cloud to the target point cloud $\small \mathbf{T}_{t,s}^{\text{guess}}$. After the iterative procedure of ICP\cite{besl1992method}, the outputs are the final transformation estimation $\small \mathbf{T}_{t,s}$ and its accuracy evaluation indexes.
	
	 \subsubsection{\textbf{\small{Multi-class closest point association}}}
	 For each iteration, the closest point correspondences are determined by nearest neighbor (NN) searching within each feature point category ($\mathcal{G}$, $\mathcal{F}$, $\mathcal{R}$, $\mathcal{P}$, $\mathcal{B}$, $\mathcal{V}$) under an ever-decreasing distance threshold. Direction consistency check of the normal vector $\boldsymbol{n}$ and the primary vector $\boldsymbol{v}$ are further applied on the planar points ($\mathcal{G}$, $\mathcal{F}$, $\mathcal{R}$) and the linear points ($\mathcal{P}$, $\mathcal{B}$) respectively. Constraints on category, distance and direction make the point association more robust and efficient.
	
	\subsubsection{\textbf{\small{Multi-metric transformation estimation}}}
	
	 Suppose $\mathbf{q}_i,\mathbf{p}_i$ are corresponding points in the source point cloud and the target point cloud, the residual $\mathbf{r}_i$ after certain rotation $\mathbf{R}$ and translation $\mathbf{t}$ of the source point cloud is defined as:
\begin{equation}
\footnotesize
  \mathbf{r}_i=\mathbf{q}_i-\mathbf{p'}_i=\mathbf{q}_i-\left( \mathbf{Rp}_i+\mathbf{t} \right).
  \label{eq:residual_vector}
\end{equation}

\begin{figure}[!t]
\centering
\setlength{\abovecaptionskip}{-2pt}
\subfigure[point to point]{
\begin{minipage}[t]{0.3\linewidth}
\centering
\includegraphics[width=1in]{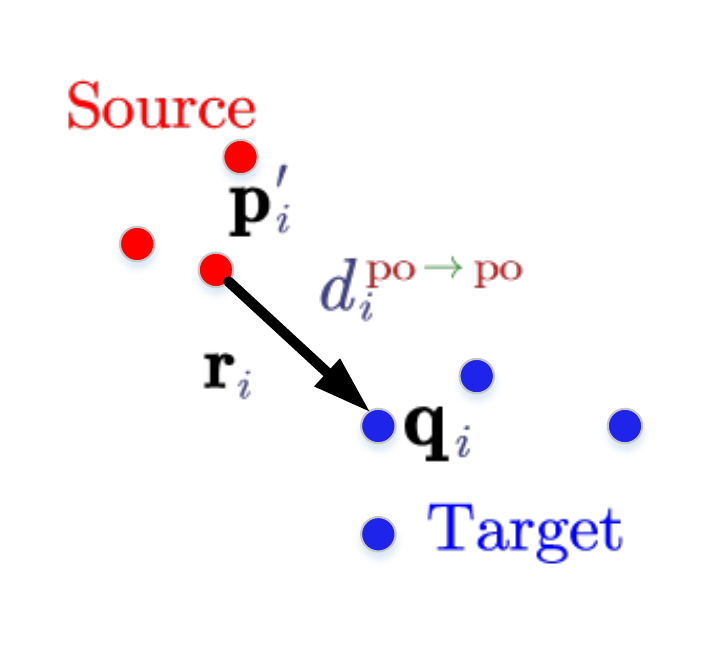}
\end{minipage}%
}%
\subfigure[point to plane]{
\begin{minipage}[t]{0.3\linewidth}
\centering
\includegraphics[width=1in]{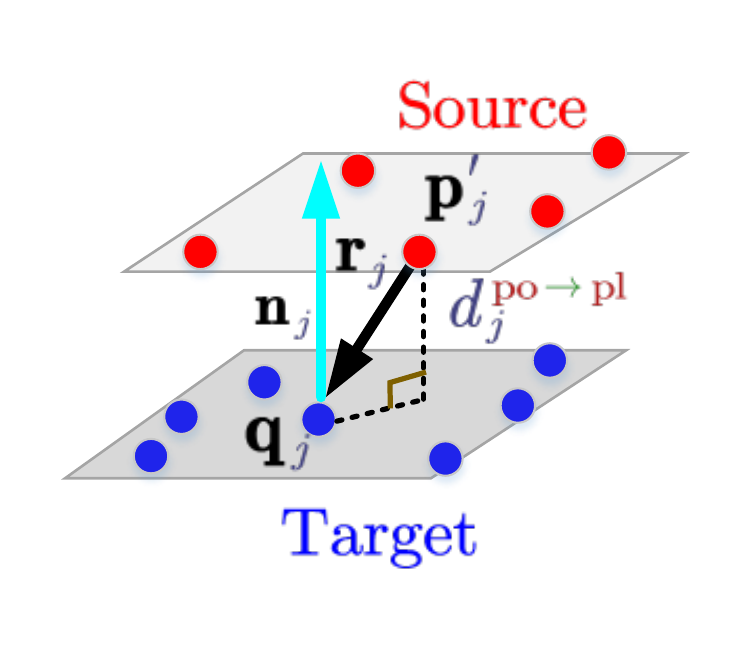}
\end{minipage}%
}%
\subfigure[point to line]{
\begin{minipage}[t]{0.3\linewidth}
\centering
\includegraphics[width=1in]{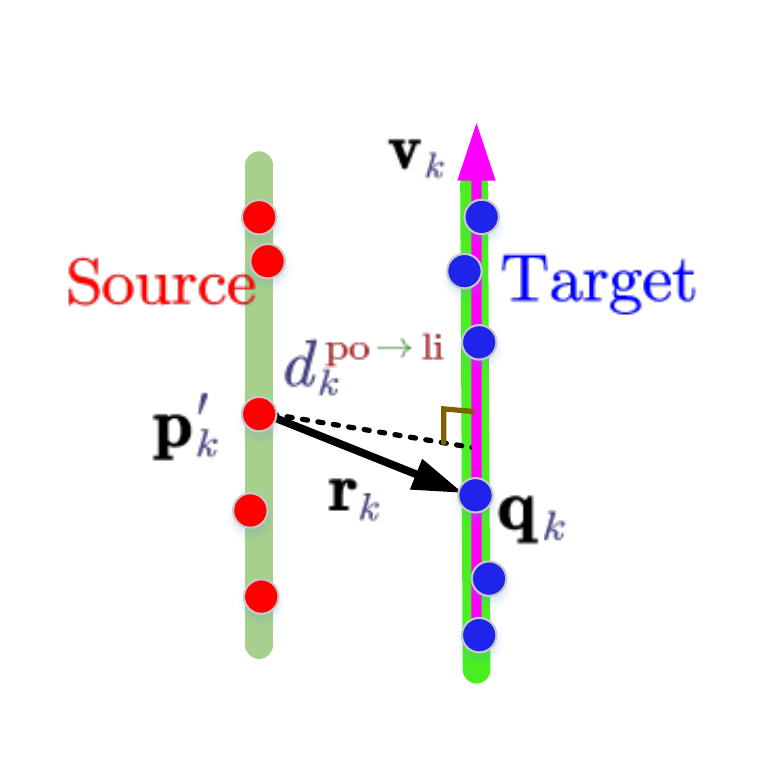}
\end{minipage}
}%
\centering
\caption{Overview of three different distance error metrics}
\label{fig:metrics}
\vspace{-16pt}
\end{figure}

With the multi-class correspondences, we expect to estimate the optimal transformation $\{\mathbf{R}^*,\mathbf{t}^*\}$ that jointly minimized the point-to-point, point-to-plane and point-to-line distance between the vertex $\mathcal{V}$, planar ($\mathcal{G}$, $\mathcal{F}$, $\mathcal{R}$) and linear ($\mathcal{P}$, $\mathcal{B}$) point correspondences.
As shown in Fig.~\ref{fig:metrics}, the point-to-point (plane, line) distance can be calculated as
$\small d_{i}^{\text{po}\rightarrow \text{po}}=\lVert \mathbf{r}_i \rVert _2$, $\small d_{j}^{\text{po}\rightarrow \text{pl}}=\mathbf{r}_j\cdot \mathbf{n}_j$, and $\small d_{k}^{\text{po}\rightarrow \text{li}}=\lVert \mathbf{r}_k\times \mathbf{v}_k \rVert _2$ respectively from the residual, normal and primary vector ($\boldsymbol{r}$, $\boldsymbol{n}$, $\boldsymbol{v}$). Thus, the transformation estimation is formulated as a weighted least square optimization problem with the following target function:
\begin{equation}
\footnotesize
\begin{split}
   \{\mathbf{R}^*,\mathbf{t}^*\} =\underset{\left\{ \mathbf{R,t} \right\}}{\text{argmin}}&\sum_i{w_i\left( d_{i}^{\text{po}\rightarrow \text{po}} \right) ^2}+\sum_j{w_j\left( d_{j}^{\text{po}\rightarrow \text{pl}} \right) ^2} \\
   +&\sum_k{w_k\left( d_{k}^{\text{po}\rightarrow \text{li}} \right) ^2}
\end{split}
\ ,
\label{eq:optimization_function_mulls}
\end{equation}

\noindent where $w$ is the weight for each correspondence. Under the tiny angle assumption, $\small\mathbf{R}$ can be approximated as:
\begin{equation}
\footnotesize
\mathbf{R}\approx \left[ \begin{matrix}
	1&		-\gamma&		\beta\\
	\gamma&		1&		-\alpha\\
	-\beta&		\alpha&		1\\
\end{matrix} \right] =\left[ \begin{array}{c}
	\alpha\\
	\beta\\
	\gamma\\
\end{array} \right] _{\times}+\mathbf{I}_{3}
\label{eq:linearR}
\ ,
\end{equation}
\noindent where $\alpha$, $\beta$ and $\gamma$ are roll, pitch and yaw, respectively under $\small x-y'-z''$ Euler angle convention. Regarding the unknown parameter vector as $\small \xi=\left[\begin{matrix}
	t_x&		t_y&		t_z&		\alpha&		\beta&		\gamma\\
\end{matrix} \right] ^\mathsf{T}$, (\ref{eq:optimization_function_mulls}) becomes a weighted linear least square problem. It is solved by Gauss-Markov parameter estimation with the functional model $\small\mathbf{e}=\mathbf{A\xi }-\mathbf{b}$ and the stochastic model $\small\text{D}\left\{ \mathbf{e} \right\} =\sigma _{0}^{2}\mathbf{P}^{-1}$, where $\sigma _{0}$ is a prior standard deviation, and $\mathbf{P}=\text{diag}\left({w_1},\cdots ,w_n \right)$ is the weight matrix. Deduced from (\ref{eq:optimization_function_mulls}), the overall design matrix $\small \mathbf{A} \in \mathbb{R}^{n\times 6}$ and observation vector $\small \mathbf{b} \in \mathbb{R}^{n\times 1}$ is formulated as:
\begin{equation}
	\footnotesize
	\begin{split}
		\mathbf{A}=&\left[\begin{matrix}
			\mathbf{A}_{i}^{\text{po}\rightarrow \text{po}\mathsf{T}}&\cdots& \mathbf{A}_{j}^{\text{po}\rightarrow \text{pl}\mathsf{T}}&\cdots&	 \mathbf{A}_{k}^{\text{po}\rightarrow \text{li}\mathsf{T}}&\cdots\\
		\end{matrix}\right]^\mathsf{T}
		\\
		\mathbf{b}=&\left[\begin{matrix}
			\mathbf{b}_{i}^{\text{po}\rightarrow \text{po}\mathsf{T}}&\cdots&\mathbf{b}_{j}^{\text{po}\rightarrow \text{pl}\mathsf{T}}&\cdots&		\mathbf{b}_{k}^{\text{po}\rightarrow \text{li}\mathsf{T}}&\cdots\\
		\end{matrix}\right]^\mathsf{T}
	\end{split}
	\label{eq:Ab_all}
	\ ,
\end{equation}
\noindent where the components of $\mathbf{A}$ and $\mathbf{b}$ for each correspondence under point-to-point (plane, line) error metric are defined as:
\begin{equation}
	\footnotesize 
	\begin{aligned}
        &\mathbf{A}_{i}^{\text{po}\rightarrow \text{po}}=\left[
		\mathbf{I}_3
		\ \left[
		\mathbf{p}_i \right] _{\times}
		\right]
		&&
		\mathbf{b}_{i}^{\text{po}\rightarrow \text{po}}=\mathbf{q}_i-\mathbf{p}_i
		\\ 
	    & \mathbf{A}_{j}^{\text{po}\rightarrow \text{pl}}=
	    \left[ 
	    \mathbf{n}_{j}^\mathsf{T}
	    \
		\left(\mathbf{p}_j\times \mathbf{n}_j \right) ^\mathsf{T}
		\right]
		&&
		\mathbf{b}_{j}^{\text{po}\rightarrow \text{pl}}=\mathbf{n}_{j}^\mathsf{T}\left( \mathbf{q}_j-\mathbf{p}_j \right)
		\\
		& \mathbf{A}_{k}^{\text{po}\rightarrow \text{li}}=\left[
		\left[ \mathbf{v}_k \right] _{\times}  \ 
		\left( \mathbf{v}_{k}^\mathsf{T}\mathbf{p}_k \right) \mathbf{I}_{\mathbf{3}}-\mathbf{p}_k\mathbf{v}_{k}^\mathsf{T} \right]
		&&
		\mathbf{b}_{k}^{\text{po}\rightarrow \text{li}}=\mathbf{v}_k\times \left( \mathbf{q}_k-\mathbf{p}_k \right)
	\end{aligned}
	\label{eq:Aandb}
	\ .
\end{equation}
Therefore, the estimation of unknown vector $\hat{\mathbf{\xi }}$ is:
\begin{equation}
\footnotesize
\begin{split}
\hat{\mathbf{\xi }}=&\underset{\mathbf{\xi }}{\text{argmin}}\left( \mathbf{A\xi }-\mathbf{b} \right) ^\mathsf{T}\mathbf{P}\left( \mathbf{A\xi }-\mathbf{b} \right) =\left( \mathbf{A}^\mathsf{T}\mathbf{PA} \right) ^{-1}\left( \mathbf{A}^\mathsf{T}\mathbf{Pb} \right) 
\\
=&\left( \sum_{i=1}^n{\mathbf{A}_{i}^\mathsf{T}w_i\mathbf{A}_{\text{i}}} \right) ^{-1}\sum_{i=1}^n{\mathbf{A}_{i}^\mathsf{T}w_i\mathbf{b}_i}
\end{split}
\label{eq:unknown_estimation}
\ ,
\end{equation}
\noindent where both $\footnotesize\mathbf{A}^\mathsf{T}\mathbf{PA}$ and $\footnotesize\mathbf{A}^\mathsf{T}\mathbf{Pb}$ can be accumulated efficiently from each correspondence in parallel. Finally, the transformation matrix $\small\hat{\mathbf{T}}\{\hat{\mathbf{R}},\hat{\mathbf{t}}\}$ is reconstructed from $\hat{\mathbf{\xi }}$.

\subsubsection{\textbf{\small{Multi-strategy weighting function}}}
	Throughout the iterative procedure, a multi-strategy weighting function based on residuals, balanced direction contribution and intensity consistency is proposed. The weight $w_i$ for each correspondence is defined as $\small w_i=w_{i}^{\left( \text{residual} \right)}\times w_{i}^{\left( \text{balanced} \right)}\times w_{i}^{\left( \text{intensity} \right)}
	$, whose multipliers are explained as follows.

First, to better deal with the outliers, we present a residual weighting function deduced from the general robust kernel function covering a family of M-estimators\cite{barron2019general}:
\begin{equation}
\footnotesize
w_{i}^{\left( \text{residual} \right)}=
\begin{cases}
\begin{aligned}
	&1, &&\text{if\,}\kappa =2\\
	&\frac{2\epsilon _i}{\epsilon _{i}^{2}+2},&&\text{if\,}\kappa =0\\
	&\epsilon _i\left( \frac{\epsilon _{i}^{2}}{\left| \kappa -2 \right|}+1 \right) ^{\frac{\kappa}{2}-1},&&\text{otherwise}\\
\end{aligned}
\end{cases}
\ ,
\label{eq:w_res}
\end{equation}
\noindent where $\kappa$ is the coefficient for the kernel's shape, $\epsilon _i=d_i/\delta$ is the normalized residual and $\delta$ is the inlier noise threshold. Although an adaptive searching for the best $\kappa$ is feasible \cite{chebrolu2020adaptive}, it would be time-consuming. Therefore, we fix $\kappa=1$ in practice, thus leading to a pseudo-Huber kernel function.

Second, the contribution of the correspondences is not always balanced in x, y, z direction. 
The following weighting function considering the number of correspondences from each category is proposed to guarantee observability.  
\begin{equation}
\footnotesize
w_{i}^{\left( \text{balanced} \right)}=\begin{cases}
    \begin{aligned}
	&\frac{\left| \mathcal{F} \right|+2\left| \mathcal{P} \right|-\left| \mathcal{B} \right|}{2\left( \left| \mathcal{G} \right|+\left| \mathcal{R} \right| \right)},&& i\in \mathcal{G}\ \text{or\ }i\in \mathcal{R}\\\
	&1,&&\text{otherwise}\\
	\end{aligned}
\end{cases}
.
\label{eq:bal}
\end{equation}

Third, since the intensity channel provides extra information for registration, (\ref{eq:w_int}) is devised to penalize the contribution of correspondences with large intensity inconsistency.
\begin{equation}
\footnotesize
w_{i}^{\left( \text{intensity} \right)}=e^{-\frac{\left| \varDelta I_{\text{i}} \right|}{I_{\max}}}
\ .
\label{eq:w_int}
\end{equation}

	\subsubsection{\textbf{\small{Multi-index registration quality evaluation}}}
	
After the convergence of ICP, the posterior standard deviation $\hat{\sigma}$ and information matrix $\mathcal{I}$ of the registration are calculated as:
\begin{equation}
\footnotesize
    \hat{\sigma}^2=\frac{1}{n-6}\left( \mathbf{A\hat{\xi}}-\mathbf{b} \right) ^\mathsf{T}\mathbf{P}\left( \mathbf{A\hat{\xi}}-\mathbf{b} \right),  \mathcal{I}=\frac{1}{\hat{\sigma}^2}(\mathbf{A}^\mathsf{T}\mathbf{PA})
    \label{eq:info_mat}
    \ .
\end{equation}	

\noindent 
\noindent where $\hat{\sigma}$, $\mathcal{I}$ together with the nonground point overlapping ratio $O_{\text{ts}}$, are used for evaluating the registration quality.

\subsection{MULLS front-end}
\label{sec:MULLS_front}

	\begin{figure}[!t]
		\centering
	\setlength{\abovecaptionskip}{-2pt}
		\includegraphics[width=0.45\textwidth]{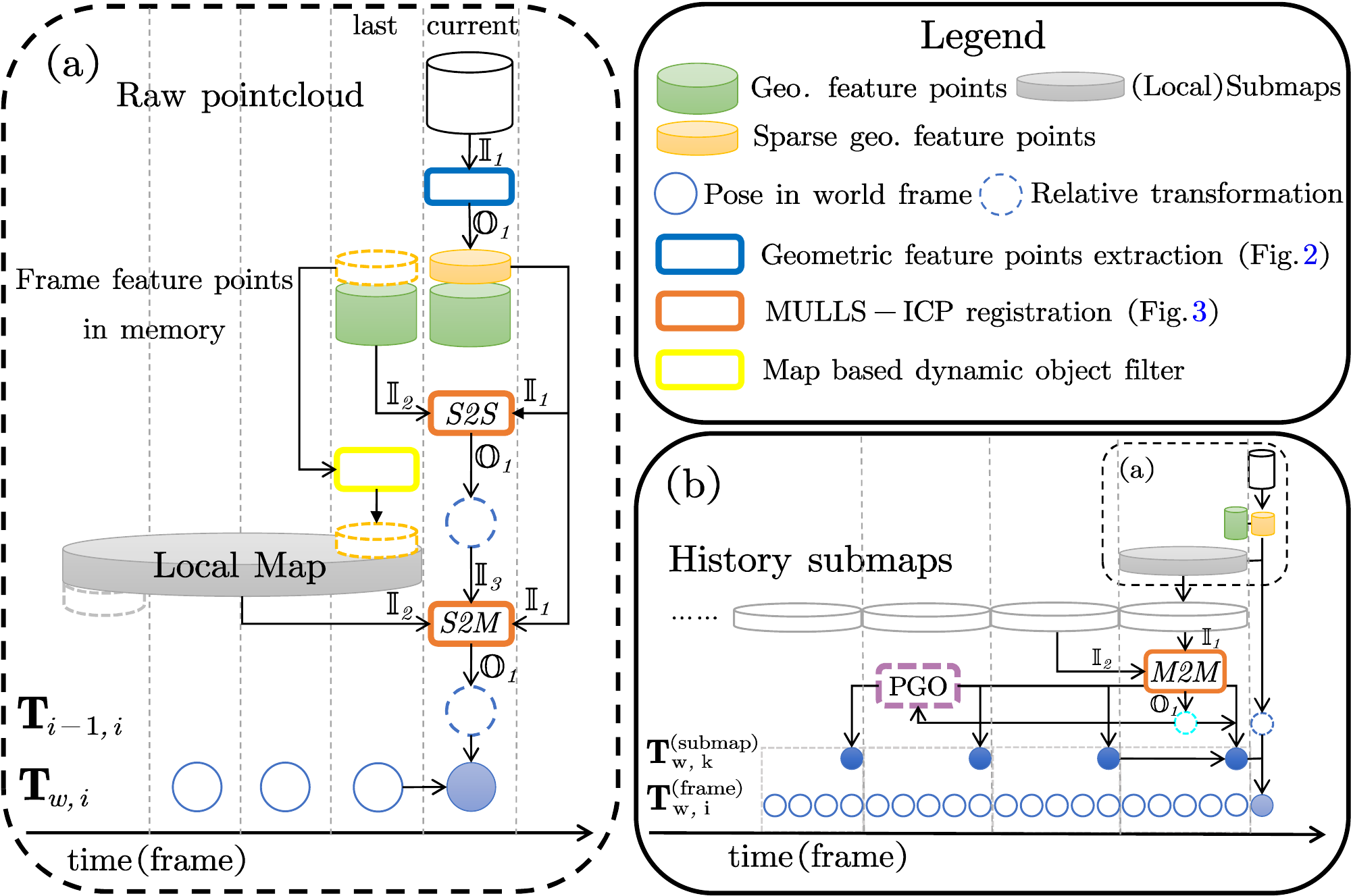}
		\caption{Overall workflow of MULLS-SLAM: (a) front-end, (b) back-end.}
		\label{fig:workflow}
	\vspace{-5pt}
	\end{figure}
	
As shown in Fig.~\ref{fig:workflow}(a), the front-end of MULLS is a combination of \ref{sec:geoextract}, \ref{sec:mullsicp} and a map management module. With an incoming scan, roughly classified feature points are extracted and further downsampled into a sparser group of points for efficiency. Besides, a local map containing static feature points from historical frames is maintained with the reference pose of the last frame. 
Taking the last frame's ego-motion as an initial guess, the scan-to-scan MULLS-ICP is conducted between the sparser feature points of the current frame and the denser feature points of the last frame with only a few iterations. The estimated transformation is provided as a better initial guess for the following scan-to-map MULLS-ICP. It regards the feature points in the local map as the target point cloud and keeps iterating until the convergence ($\hat{\mathbf{\xi }}$ smaller than a threshold). The sparser group of feature points are appended to the local map after the map-based dynamic object filtering. For this step, within a distance threshold to the scanner, those nonground feature points far away from their nearest neighbors with the same category in the local map are filtered.
The local map is then cropped with a fixed radius. 
Only the denser group of feature points of the current frame are kept for the next epoch.



\subsection{MULLS back-end}
\label{sec:backend}
\begin{figure}[!t]
		\centering
		\setlength{\abovecaptionskip}{-2pt}
		\includegraphics[width=0.42\textwidth]{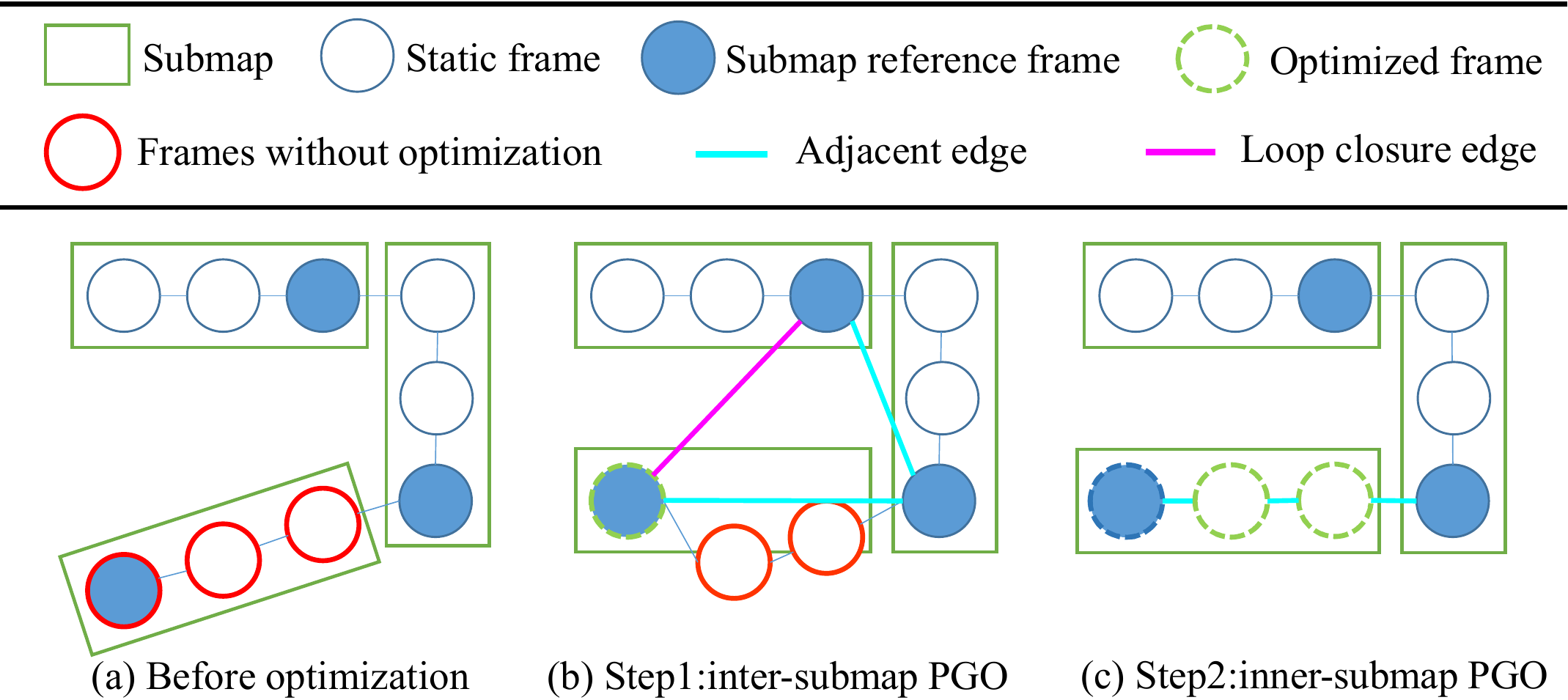}
		\caption{Inter-submap and inner-submap pose graph optimization}
		\label{fig:pgo}
     	\vspace{-10pt}
	\end{figure}

 As shown in Fig.~\ref{fig:workflow}(b), the periodically saved submap is the processing unit. The adjacent and loop closure edges among the submaps are constructed and verified by the certificated and efficient TEASER\cite{Yang20tro-teaser} global registration. Its initial correspondences are determined according to the cosine similarity of NCC features encoded in \ref{sec:geoextract} among the vertex keypoints. Taking TEASER's estimation as the initial guess, the map-to-map MULLS-ICP is used to refine inter-submap edges with accurate transformation and information matrices calculated in \ref{sec:mullsicp}. Those edges with higher $\hat{\sigma}$ or lower $O_{\text{ts}}$ than thresholds would be deleted. 
  As shown in Fig.~\ref{fig:pgo}, once a loop closure edge is added, the pose correction of free submap nodes is achieved by the inter-submap PGO. Hierarchically, the inner-submap PGO fixes each submap's reference frame and adjusts the other frames' pose. 
  
	\section{Experiments}
	\label{sec:experi}
	

	
	
	
	


The proposed MULLS system is evaluated qualitatively and quantitatively on KITTI, MIMAP and HESAI datasets, covering various outdoor and indoor scenes using seven different types of LiDAR, as summarized in Table \textcolor{blue}{I}. All the experiments are conducted on a PC equipped with an Intel Core i7-7700HQ@2.80GHz CPU for fair comparison.

\begin{table}[!b]
\scriptsize
\vspace{-4pt}
\centering
\label{tab:datasets}
\setlength{\tabcolsep}{0.5mm}{
\begin{tabular}{ccccc}
\multicolumn{5}{c}{\footnotesize TABLE I: The proposed method is evaluated on three datasets} \\[0pt]
\multicolumn{5}{c}{\footnotesize  collected by various LiDARs in various scenarios.} \\[0pt]
\hline
Dataset & LiDAR                             & Scenario                            & \#Seq.~ & \#Frame \\ \hline
KITTI\cite{geiger2012we}   & HDL64E                             & urban, highway, country    & 22         & 43k     \\
MIMAP\cite{wang2020isprs}   & VLP32C, HDL32E                       & indoor                              & 3          & 35k     \\
HESAI   & PandarQTLite, XT, 64, 128 & residential, urban, indoor & 8          & 30k     \\ \hline
\end{tabular}}
\end{table}

\subsection{Quantitative experiment}

	\begin{figure}[!t]
		\centering
		\setlength{\abovecaptionskip}{-16pt}
		\includegraphics[width=0.48\textwidth]{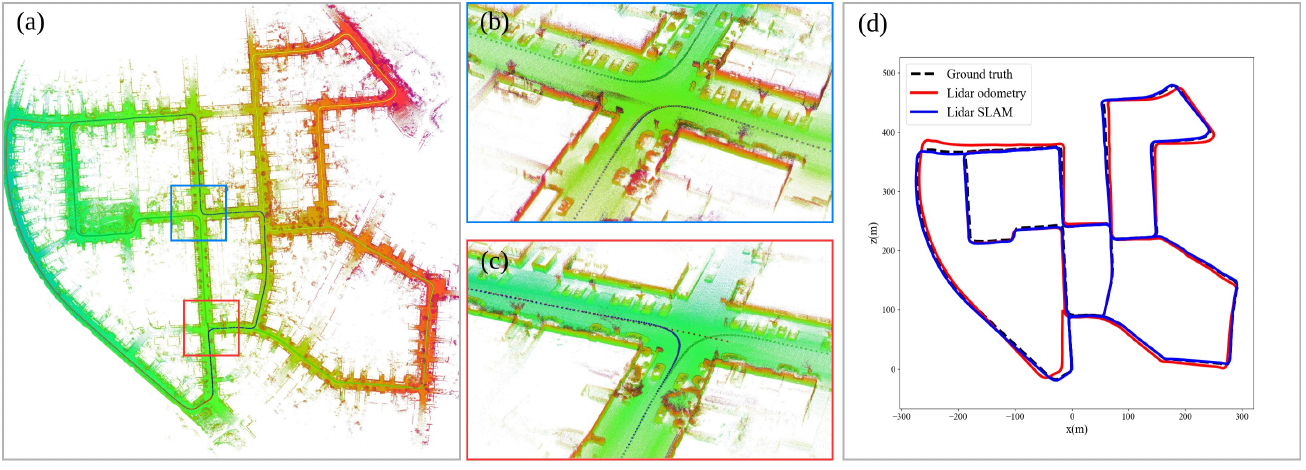}
		\caption{MULLS's result on urban scenario (KITTI seq.00): (a) overview, (b,c) map in detail of loop closure areas, (d) trajectory comparison.}
		\label{fig:kitti00}
		\vspace{-8pt}
	\end{figure}
	
	\begin{figure}[!t]
		\centering
    \setlength{\abovecaptionskip}{-16pt}
		\includegraphics[width=0.48\textwidth]{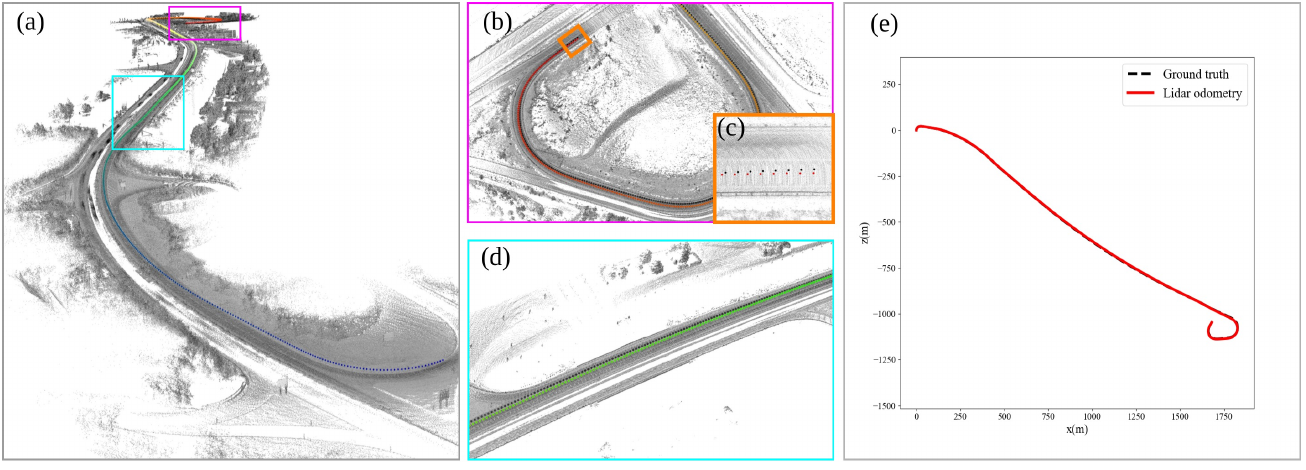}
		\caption{MULLS's result on highway scenarios (KITTI seq.01): (a) overview, (b) map and trajectory (black: ground truth) at the end of the sequence, (c) lane-level final drift, (d) map and trajectory of a challenging scene with high similarity and few features, (e) trajectory comparison (almost overlapped).}
		\label{fig:kitti01}
    \vspace{-10pt}
	\end{figure}

The quantitative evaluations are done on the KITTI Odometry/SLAM dataset \cite{geiger2012we}, which is composed of 22 sequences of data with more than 43k scans captured by a Velodyne HDL-64E LiDAR. KITTI covers various outdoor scenarios such as urban road, country road, and highway. Seq.~0-10 are provided with GNSS-INS ground truth pose while seq.~11-21 are used as the test set for online evaluation and leaderboard without provided ground truth. For better performance, an intrinsic angle correction of $0.2^\circ$ is applied\cite{deschaud2018imls,yin2020caelo}. Following the odometry evaluation criterion in~\cite{geiger2012we}, the average translation \& rotation error (ATE \& ARE) are used for localization accuracy evaluation. 
The performance of the proposed MULLS-LO (w/o loop closure) and MULLS-SLAM (with loop closure), as well as 10 state-of-the-art LiDAR SLAM solutions (results taken from their original papers and KITTI leaderboard), are reported in Table \textcolor{blue}{II}. MULLS-LO achieves the best ATE (0.49\%) on seq.~00-10 
and runner-up on online test sets. With loop closure and PGO, MULLS-SLAM gets the best ARE (0.13$^\circ$/100m) but worse ATE. This is because ATE \& ARE can't fully reflect the gain of global positioning accuracy as they are measured locally within a distance of 800m. 
Besides, the maps and trajectories constructed from MULLS for two representative sequences are shown in Fig.~\ref{fig:kitti00} and \ref{fig:kitti01}. 
It has only a lane-level drift in the end on the featureless and highly dynamic highway sequence, as shown in Fig.~\ref{fig:kitti01}(c).

Some variants of MULLS are also compared in Table \textcolor{blue}{II}. It is shown that scan-to-map registration can boost accuracy a lot. With only one iteration, MULLS\textit{(m1)} already ranks 5th among the listed methods.  With five iterations, MULLS\textit{(m5)} get ATE of about 0.6\%, which is close to the converged MULLS (with about 15 iterations on average)'s performance. The results indicate that we can sacrifice a bit of accuracy for faster operation on platforms with limited computing power. Besides, scan-to-scan registration before scan-to-map registration is not necessary once the local map is constructed as \textit{(s5m5)} has similar performance to \textit{(m5)}. We thus use scan-to-map registration till convergence \textit{(mc)} in MULLS.    
  
\begin{table*}[]
\scriptsize
\renewcommand\arraystretch{1.1}
\begin{center}
\setlength{\tabcolsep}{0.6mm}{
\begin{tabular}{|c|lllllllllllll|c|}  
\multicolumn{15}{c}{\footnotesize TABLE II: Quantitative evaluation and comparison on KITTI dataset.} \\[0pt]
\multicolumn{15}{l}{All errors are represented as \textbf{ATE[\%] / ARE[$^\circ$/100m]} (the smaller the better). \textcolor{red}{\textbf{Red}} and \textcolor{blue}{\textbf{blue}} fonts denote the first and second place, respectively.}\\[0pt]
\multicolumn{15}{l}{\textbf{Denotations}: U: urban road, H: highway, C: country road, *: with loop closure, -: not available, \textit{\textbf{(mc)}}:scan-to-\textbf{m}ap registration till \textbf{c}onvergence, \textit{\textbf{(s1)}}:scan-to-\textbf{s}can registration }\\[0pt]
\multicolumn{15}{l}{ for \textbf{1} iteration, \textit{\textbf{(m5)}}:scan-to-\textbf{m}ap registration for \textbf{5} iterations, \textit{\textbf{(s5m5)}}:scan-to-\textbf{s}can for \textbf{5} iterations before scan-to-\textbf{m}ap for \textbf{5} iterations}\\[0pt]
\hline
  Method & 00~U* & 01~H & 02~C* & 03~C & 04~C & 05~C* & 06~U* & 07~U* & 08~U* & 09~C* & 10~C & 00-10\tiny{mean}& 11-21\tiny{mean} & time(s)/frame\\ 
 \hline
 \hline
  
 LOAM~\cite{zhang2014loam} & 0.78\textbf{/} - & 1.43\textbf{/} - & 0.92\textbf{/} - & 0.86\textbf{/} - & 0.71\textbf{/} - & 0.57\textbf{/} - & 0.65\textbf{/} - & 0.63\textbf{/} - & 1.12\textbf{/} - & 0.77\textbf{/} - & 0.79\textbf{/} - & 0.84\textbf{/} - & \textcolor{red}{\textbf{0.55}}\textbf{/}\textcolor{red}{\textbf{0.13}} & 0.10 \\
 \hline
 
 IMLS-SLAM~\cite{deschaud2018imls} & \textcolor{red}{\textbf{0.50}}\textbf{/} - & 0.82\textbf{/} - & \textcolor{red}{\textbf{0.53}}\textbf{/} - & 0.68\textbf{/} - & \textcolor{red}{\textbf{0.33}}\textbf{/} - & 0.32\textbf{/} - & 0.33\textbf{/} - & 0.33\textbf{/} - & \textcolor{red}{\textbf{0.80}}\textbf{/} - & 0.55\textbf{/} - & \textcolor{blue}{\textbf{0.53}}\textbf{/} - & \textcolor{blue}{\textbf{0.52}}\textbf{/} - & 0.69\textbf{/}0.18 & 1.25\\
 \hline
 
 MC2SLAM~\cite{neuhaus2018mc2slam} & \textcolor{blue}{\textbf{0.51}}\textbf{/} - & \textcolor{blue}{\textbf{0.79}}\textbf{/} - & \textcolor{blue}{\textbf{0.54}}\textbf{/} - & \textcolor{blue}{\textbf{0.65}}\textbf{/} - &
 0.44\textbf{/} - & \textcolor{red}{\textbf{0.27}}\textbf{/} - & 0.31\textbf{/} - & 0.34\textbf{/} - & 0.84\textbf{/} - & \textcolor{red}{\textbf{0.46}}\textbf{/} - &
 \textcolor{red}{\textbf{0.52}}\textbf{/} - & \textcolor{blue}{\textbf{0.52}}\textbf{/} - & 0.69\textbf{/}\textcolor{blue}{\textbf{0.16}} & 0.10\\
 \hline
 
 S4-SLAM~\cite{zhou2020s4}* & 0.62\textbf{/} - & 1.11\textbf{/} - & 1.63\textbf{/} - & 0.82\textbf{/} - & 0.95\textbf{/} - & 0.50\textbf{/} - & 0.65\textbf{/} - & 0.60\textbf{/} - & 1.33\textbf{/} - & 1.05\textbf{/} - & 0.96\textbf{/} - & 0.92\textbf{/} - & 0.93\textbf{/}0.38 & 0.20\\
 \hline
 
 PSF-LO~\cite{chen2020psf} & 0.64\textbf{/} - & 1.32\textbf{/} - & 0.87\textbf{/} - & 0.75\textbf{/} - & 0.66\textbf{/} - & 0.45\textbf{/} - & 0.47\textbf{/} - & 0.46\textbf{/} - & 0.94\textbf{/} - & 0.56\textbf{/} - & 0.54\textbf{/} - & 0.74\textbf{/} - & 0.82\textbf{/}0.32 & 0.20\\
 \hline
 
 SUMA$^{++}$~\cite{chen2019suma++}*& 0.64\textbf{/}0.22 & 1.60\textbf{/}0.46 &1.00\textbf{/}0.37 & 0.67\textbf{/}0.46 & 0.37\textbf{/}\textcolor{blue}{\textbf{0.26}} & 0.40\textbf{/}0.20 & 0.46\textbf{/}0.21 & 0.34\textbf{/}0.19 & 1.10\textbf{/}0.35 & \textcolor{blue}{\textbf{0.47}}\textbf{/}0.23 & 0.66\textbf{/}0.28 & 0.70\textbf{/}0.29 & 1.06\textbf{/}0.34 & 0.10\\
 \hline
 
 LiTAMIN2~\cite{yokozuka2021litamin2}* & 0.70\textbf{/}0.28 & 2.10\textbf{/}0.46 & 0.98\textbf{/}0.32 & 0.96\textbf{/}0.48 & 1.05\textbf{/}0.52 & 0.45\textbf{/}0.25 & 0.59\textbf{/}0.34 & 0.44\textbf{/}0.32 & 0.95\textbf{/}0.29 & 0.69\textbf{/}0.40 & 0.80\textbf{/}0.47 & 0.85\textbf{/}0.33 & \quad - \textbf{/} - & \textbf{0.01} \\
 \hline
 
 LO-Net~\cite{li2019net} & 0.78\textbf{/}0.42 & 1.42\textbf{/}0.40 & 1.01\textbf{/}0.45 & 0.73\textbf{/}0.59 & 0.56\textbf{/}0.54 & 0.62\textbf{/}0.35 & 0.55\textbf{/}0.35 & 0.56\textbf{/}0.45 & 1.08\textbf{/}0.43 & 0.77\textbf{/}0.38 & 0.92\textbf{/}0.41 & 0.83\textbf{/}0.42 & 1.75\textbf{/}0.79 & 0.10 \\
 \hline

 FALO~\cite{garcia2020fail}& 1.28\textbf{/}0.51 & 2.36\textbf{/}1.35 & 1.15\textbf{/}0.28 & 0.93/\textcolor{blue}{\textbf{0.24}} & 0.98\textbf{/}0.33 & 0.45\textbf{/}0.18 & \quad - \textbf{/} -& 0.44\textbf{/}0.34 & \quad - \textbf{/} - & 0.64\textbf{/}\textcolor{blue}{\textbf{0.13}} & 0.83\textbf{/}\textcolor{red}{\textbf{0.17}} & 1.00\textbf{/}0.39 & \quad - \textbf{/} - & 0.10 \\
 \hline
 
 LoDoNet~\cite{zheng2020lodonet} & 1.43\textbf{/}0.69 & 0.96\textbf{/}\textcolor{blue}{\textbf{0.28}} & 1.46\textbf{/}0.57 &2.12\textbf{/}0.98 & 0.65\textbf{/}0.45 & 1.07\textbf{/}0.59 & 0.62\textbf{/}0.34 & 1.86\textbf{/}1.64 & 2.04\textbf{/}0.97 & 0.63\textbf{/}0.35 & 1.18\textbf{/}0.45 & 1.27\textbf{/}0.66 &  \quad - \textbf{/} - & - \\
 \hline

 \textbf{MULLS-LO}\textit{(mc)} & \textcolor{blue}{\textbf{0.51}}\textbf{/}\textcolor{blue}{\textbf{0.18}}  & \textcolor{red}{\textbf{0.62}}\textbf{/}\textcolor{red}{\textbf{0.09}} & 0.55\textbf{/}\textcolor{blue}{\textbf{0.17}} & \textcolor{red}{\textbf{0.61}}\textbf{/}\textcolor{red}{\textbf{0.22}} &
 \textcolor{blue}{\textbf{0.35}}\textbf{/}\textcolor{red}{\textbf{0.08}} & 
 \textcolor{blue}{\textbf{0.28}}\textbf{/}\textcolor{blue}{\textbf{0.17}} &
 \textcolor{red}{\textbf{0.24}}\textbf{/}\textcolor{blue}{\textbf{0.11}} &
 \textcolor{blue}{\textbf{0.29}}\textbf{/}\textcolor{blue}{\textbf{0.18}} &
 \textcolor{red}{\textbf{0.80}}\textbf{/}\textcolor{blue}{\textbf{0.25}} &
 0.49\textbf{/}0.15 & 0.61\textbf{/}\textcolor{blue}{\textbf{0.19}} &
 \textcolor{red}{\textbf{0.49}}\textbf{/}\textcolor{blue}{\textbf{0.16}}&
 \textcolor{blue}{\textbf{0.65}}\textbf{/}0.19 & 0.08 \\

 \textbf{MULLS-SLAM}\textit{(mc)}* & 0.54\textbf{/}\textcolor{red}{\textbf{0.13}} &\textcolor{red}{\textbf{0.62}}\textbf{/}\textcolor{red}{\textbf{0.09}}  & 0.69\textbf{/}\textcolor{red}{\textbf{0.13}} & \textcolor{red}{\textbf{0.61}}\textbf{/}\textcolor{red}{\textbf{0.22}} &
\textcolor{blue}{\textbf{0.35}}\textbf{/}\textcolor{red}{\textbf{0.08}} & 
0.29\textbf{/}\textcolor{red}{\textbf{0.07}} &
 \textcolor{blue}{\textbf{0.29}}\textbf{/}\textcolor{red}{\textbf{0.08}} &
 \textcolor{red}{\textbf{0.27}}\textbf{/}\textcolor{red}{\textbf{0.11}} &
 \textcolor{blue}{\textbf{0.83}}\textbf{/}\textcolor{red}{\textbf{0.17}} &
 0.51\textbf{/}\textcolor{red}{\textbf{0.12}} & 0.61\textbf{/}\textcolor{blue}{\textbf{0.19}} &
 \textcolor{blue}{\textbf{0.52}}\textbf{/}\textcolor{red}{\textbf{0.13}}&
 \quad - \textbf{/} - & 0.10 \\
  
 \hline
 
  MULLS-LO\textit{(s1)} & 2.36\textbf{/}1.06 & 2.76\textbf{/}0.89 & 2.81\textbf{/}0.95 & 1.26\textbf{/}0.67 & 5.72\textbf{/}1.15& 
2.19\textbf{/}1.01 &
 1.12\textbf{/}0.51 &
 1.65\textbf{/}1.27 & 2.73\textbf{/}1.19 & 2.14\textbf{/}0.96 & 3.61\textbf{/}1.65 & 2.57\textbf{/}1.03
 &
 \quad - \textbf{/} - & 0.03 \\
 
 MULLS-SLAM\textit{(m1)}* & 0.79\textbf{/}0.28 & 1.01\textbf{/}0.25 & 1.46\textbf{/}0.39 & 0.79\textbf{/}0.23 &
1.02\textbf{/}0.15 & 
0.40\textbf{/}0.14 &
 0.38\textbf{/}0.13 &
 0.48\textbf{/}0.34 &
 0.94\textbf{/}0.29 &
 0.61\textbf{/}0.22 & 0.57\textbf{/}0.29 & 0.77\textbf{/}0.25
 &
 \quad - \textbf{/} - & 0.05 \\

 MULLS-SLAM\textit{(m5)}* & 0.65\textbf{/}0.14 & 0.79\textbf{/}0.16 & 1.08\textbf{/}0.27 & 0.59\textbf{/}0.25 &
0.50\textbf{/}0.11 & 
0.32\textbf{/}0.08 &
 0.36\textbf{/}0.09 &
 0.36\textbf{/}0.24 &
 0.75\textbf{/}0.18 &
 0.58\textbf{/}0.23 & 0.68\textbf{/}0.23  & 0.60\textbf{/}0.18
 &
 \quad - \textbf{/} - & 0.07 \\
 
 MULLS-SLAM\textit{(s5m5)}* & 0.66\textbf{/}0.14 & 0.82\textbf{/}0.19 & 1.01\textbf{/}0.26 & 0.59\textbf{/}0.27 &
0.63\textbf{/}0.12 & 
0.34\textbf{/}0.08 &
 0.34\textbf{/}0.08 &
 0.40\textbf{/}0.29 &
 0.81\textbf{/}0.19 &
 0.56\textbf{/}0.20 & 0.58\textbf{/}0.20  & 0.61\textbf{/}0.18
 &
 \quad - \textbf{/} - & 0.08 \\
 
  \hline

\end{tabular}
}
\end{center}
\label{tab:kitti_ford}
\vspace{-20pt}
\end{table*}



\begin{figure}[!t]
	\centering
	\setlength{\abovecaptionskip}{-14pt}
	\includegraphics[width=0.48\textwidth]{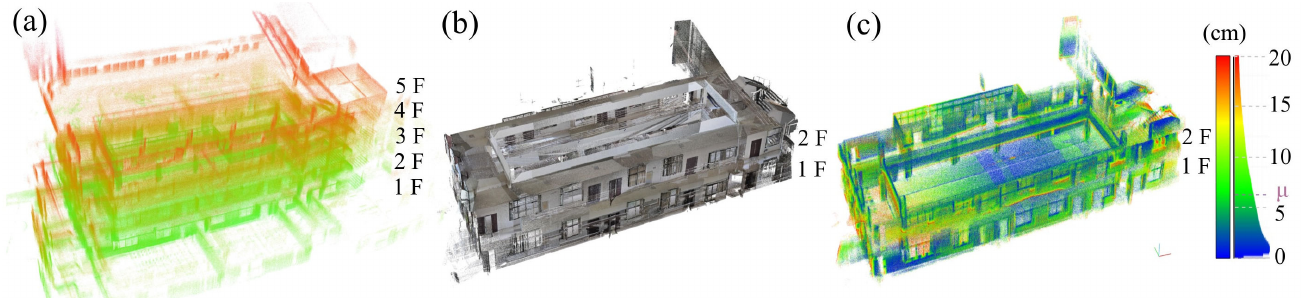}
	\caption{MULLS's result on MIMAP dataset: (a) the front view of the point cloud map of a 5-floor building generated by MULLS, (b) point cloud of the building (floor 1\&2) collected by TLS, (c) point to point distance ($\mu=6.7 \text{cm}$) between the point cloud generated from TLS and MULLS.}
	\label{fig:mimap}
     \vspace{-8pt}
\end{figure}

\begin{figure}[!t]
\centering
\setlength{\abovecaptionskip}{-14pt}
\includegraphics[width=0.48\textwidth]{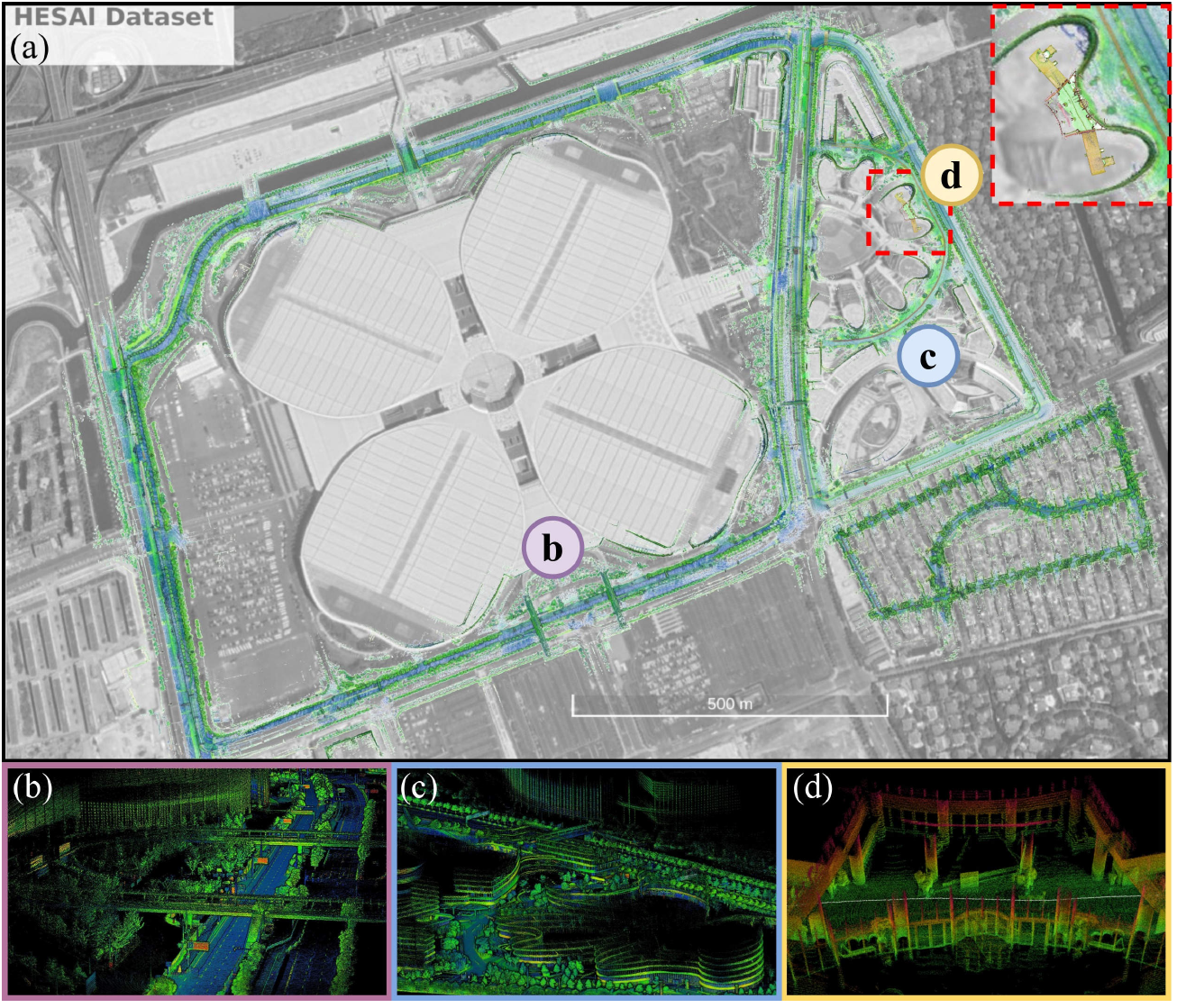}
\caption{Overview of MULLS's results on HESAI dataset: (a) MULLS SLAM map aligned with Google Earth, (b-d) example of the generated point cloud map of the urban expressway, industry park, and indoor scenario.}
\label{fig:hs}
\vspace{-16pt}
\end{figure}


\subsection{Qualitative experiments}
	
\subsubsection{\textbf{\small MIMAP}}
The ISPRS Multi-sensor Indoor Mapping and Positioning dataset\cite{wang2020isprs} is collected by a backpack mapping system with Velodyne VLP32C and HDL32E in a 5-floor building. High accuracy terrestrial laser scanning (TLS) point cloud of floor 1 \& 2 scanned by Rigel VZ-1000 is regarded as the ground truth for map quality evaluation. Fig.~\ref{fig:mimap} demonstrates MULLS's indoor mapping quality with a 6.7cm mean mapping error, which is calculated as the mean distance between each point in the map point cloud and the corresponding nearest neighbor in TLS point cloud.  

\subsubsection{\textbf{\small HESAI}}
 HESAI dataset is collected by four kinds of mechanical LiDAR (Pandar128, 64, XT, and QT Lite). Since the ground truth pose or map is not provided, the qualitative result of the consistent maps built by MULLS-SLAM indicate MULLS's ability for adapting to LiDARs with different number and distribution of beams in both the indoor and outdoor scenes, as shown in Fig.~\ref{fig:hs}.

\subsection{Ablation studies}

Furthermore, in-depth performance evaluation on the adopted categories of geometric feature points and the weighting functions are conducted to verify the proposed method's effectiveness. Table \textcolor{blue}{III} shows that vertex point-to-point correspondences undermine LO's accuracy on both sequences because points from regular structures are more reliable than those with high curvature on vegetations. Therefore, vertex points are only used as the keypoints in back-end global registration in practice. Besides, linear points (pillar and beam) are necessary for the supreme performance on highway scene. Points on guardrails and curbs are extracted as linear points to impose cross direction constraints. It also indicates MULLS may encounter problem in tunnels, where structured features are rare. As shown in Table \textcolor{blue}{IV}, all the three functions presented in \ref{sec:mullsicp} are proven effective on both sequences. Specifically, translation estimation realizes obvious improvement on featureless highway scenes with the consideration of intensity consistency. 

\begin{table}[!t] 
\scriptsize
\setlength{\abovecaptionskip}{-2pt}
\centering
\label{tab:geo-ablation}
\setlength{\tabcolsep}{1.0mm}{
\begin{tabular}{ccccccc}
\multicolumn{7}{c}{\footnotesize TABLE III: Ablation study w.r.t. geometric feature points}\\[0pt]
\multicolumn{7}{c}{( -: LiDAR odometry failed)}\\[0pt]
\hline
\multicolumn{2}{c}{$\text{po}\rightarrow\text{pl}$} & \multicolumn{2}{c}{$\text{po}\rightarrow\text{li}$} &$\text{po}\rightarrow\text{po}$& KITTI 00 U & KITTI 01 H \\
 \hline
 $\mathcal{G}$ & $\mathcal{F}$ & $\mathcal{P}$ & $\mathcal{B}$ & $\mathcal{V}$ & \multicolumn{2}{c}{ATE [\%] / ARE [$^\circ$/100m]} \\
 \hline
\cmark & \cmark & \cmark & \cmark & \xmark & 0.53 / 0.20 & \bf{0.62} / \bf{0.09} \\
\cmark & \cmark & \cmark & \xmark& \xmark & \bf{0.51}
/ \bf{0.18} & 1.02 / 0.16 \\
\cmark & \cmark & \xmark & \xmark& \xmark & 0.54
/ 0.21 & - / - \\
\cmark & \xmark & \cmark & \xmark& \xmark & 0.92
/ 0.28 & - / - \\
\cmark & \xmark & \cmark & \cmark& \xmark & 0.70
/ 0.33 & 0.68 / \bf{0.09} \\
\cmark & \cmark & \cmark & \cmark& \cmark & 0.57
/ 0.23 & 0.92 / 0.11 \\
\hline
\end{tabular}}
\vspace{-6pt}
\end{table}

\begin{table}[!t]
\scriptsize
\setlength{\abovecaptionskip}{-2pt}
\centering
\label{tab:cor-ablation}
\setlength{\tabcolsep}{1.0mm}{
\begin{tabular}{cccccc}
\multicolumn{5}{c}{\footnotesize TABLE IV: Ablation study w.r.t. weighting function}\\[0pt]
\hline
\multicolumn{3}{c}{weighting function } & KITTI 00 U & KITTI 01 H \\
\hline
 $w^{\text{(bal.)}}$ & $w^{\text{(res.)}}$ & $w^{\text{(int.)}}$ & \multicolumn{2}{c}{ATE [\%] / ARE [$^\circ$/100m]}  \\
\hline
\cmark & \cmark & \cmark & \bf{0.51}
/ \bf{0.18} & \bf{0.62} / \bf{0.09} \\
\xmark & \xmark & \xmark & 0.57
/ 0.23 & 0.87 / 0.13\\
\xmark & \cmark & \cmark  & 0.52
/ 0.19 & 0.66 / 0.10 \\
\cmark & \xmark & \cmark  & 0.54
/ 0.22 & 0.76 / 0.12 \\
\cmark & \cmark & \xmark  & 0.53
/ 0.20 & 0.83 / 0.11 \\
\hline
\end{tabular}}
\vspace{-16pt}
\end{table}

\subsection{Runtime analysis}
\begin{figure}[!t]
		\centering
	\setlength{\abovecaptionskip}{-4pt}
		\includegraphics[width=0.47\textwidth]{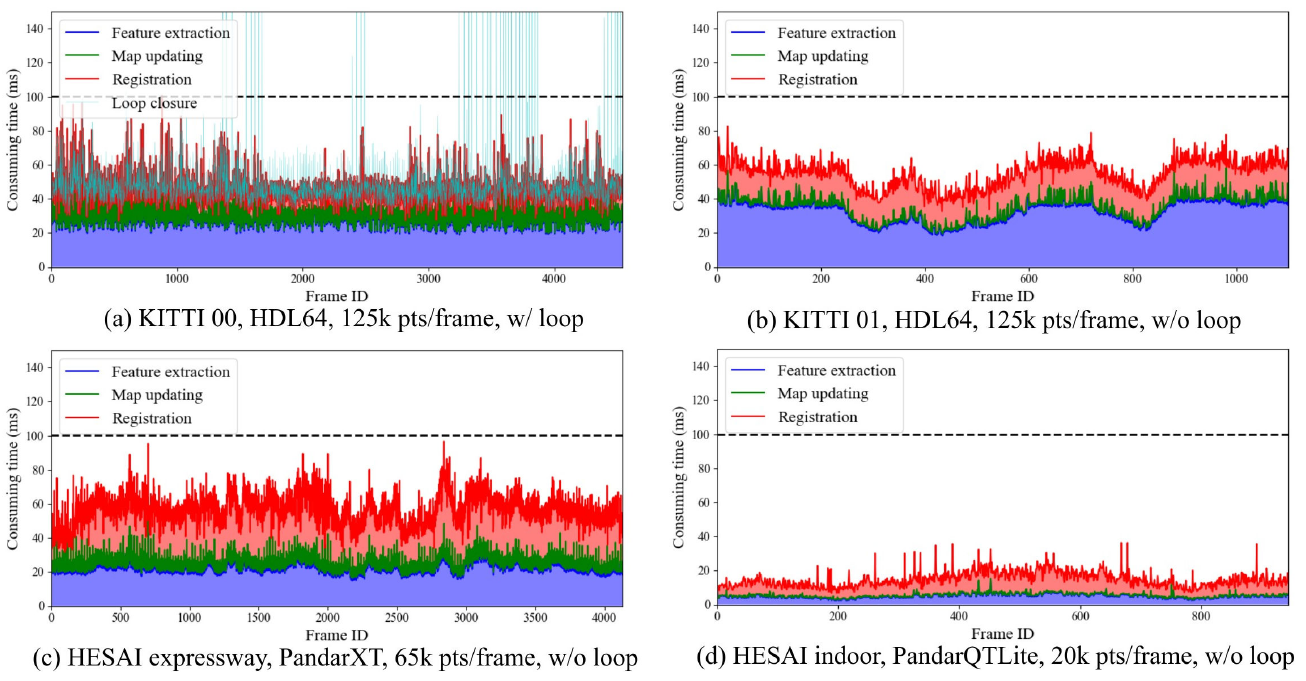}
		\caption{Run time analysis of the proposed MULLS using different types of LiDAR with or without loop closure}
		\label{fig:timing}
\vspace{-18pt}
\end{figure}

A detailed runtime analysis for each frame is shown in Table V. MULLS transformation estimation only costs 0.2 ms per ICP iteration with about 2k and 20k feature points in the source and target point cloud, respectively. Moreover, Fig.~\ref{fig:timing} shows MULLS's timing report of four typical sequences from KITTI and HESAI dataset with different point number per frame. MULLS operates faster than 10Hz on average for all the sequences, though it sometimes runs beyond 100ms on sequences with lots of loops such as Fig.~\ref{fig:timing}(a). Since the loop closure can be implemented on another thread, MULLS is guaranteed to perform in real-time on a moderate PC.



\begin{table}[]
\scriptsize
\setlength{\tabcolsep}{1.0mm}{
\begin{tabular}{|c|c|c|c|c|c|}
\multicolumn{6}{c}{\footnotesize TABLE V: Runtime analysis per frame in detail (unit: ms) }\\[1pt]
\multicolumn{6}{c}{(Typically, there're $\approx$ 2k and 20k feature points in current frame and local map)}\\[1pt]
\hline
Module & Submodule & $t_{submod}$  & \#Iter.  & $t_{mod}$ & $T$ \\ \hline
\multirow{3}{*}{Feature extraction} & ground points            & 8   & \multirow{3}{*}{1}  & \multirow{3}{*}{29} & \multirow{7}{*}{80} \\ \cline{2-3}
                                    & non-ground feature points & 20  &                     &                     &                     \\ \cline{2-3}
                                    & NCC encoding              & 1   &                     &                     &                     \\ \cline{1-5}
\multirow{2}{*}{Map updating}       & dynamic object removal    & 2   & \multirow{2}{*}{1}  & \multirow{2}{*}{3}  &                     \\ \cline{2-3}
                                    & local map updating                    & 1   &                     &                     &                     \\ \cline{1-5}
\multirow{2}{*}{Registration}       & closest point association & 3   & \multirow{2}{*}{15} & \multirow{2}{*}{48} &                     \\ \cline{2-3}
                                    & transform estimation      & 0.2 &                     &                     &                     \\ \hline
\end{tabular}}
\vspace{-10pt}
\end{table}

	\section{Conclusions}
	\label{sec:conclusion}
	In this work, we present a versatile LiDAR-only SLAM system named MULLS, which is based on the efficient multi-metric linear least square ICP. Experiments on more than 100 thousand frames collected by 7 different LiDARs show that MULLS achieves real-time performance with low drift and high map quality on various challenging outdoor and indoor scenarios regardless of the LiDAR specifications. 

	\bibliographystyle{IEEEtran}
	\bibliography{main}

\begin{thebibliography}{10}
\providecommand{\url}[1]{#1}
\csname url@rmstyle\endcsname
\providecommand{\newblock}{\relax}
\providecommand{\bibinfo}[2]{#2}
\providecommand\BIBentrySTDinterwordspacing{\spaceskip=0pt\relax}
\providecommand\BIBentryALTinterwordstretchfactor{4}
\providecommand\BIBentryALTinterwordspacing{\spaceskip=\fontdimen2\font plus
\BIBentryALTinterwordstretchfactor\fontdimen3\font minus
  \fontdimen4\font\relax}
\providecommand\BIBforeignlanguage[2]{{%
\expandafter\ifx\csname l@#1\endcsname\relax
\typeout{** WARNING: IEEEtran.bst: No hyphenation pattern has been}%
\typeout{** loaded for the language `#1'. Using the pattern for}%
\typeout{** the default language instead.}%
\else
\language=\csname l@#1\endcsname
\fi
#2}}

\bibitem{temeltas2008slam}
H.~Temeltas and D.~Kayak, ``{SLAM for robot navigation},'' \emph{IEEE Aerospace
  and Electronic Systems Magazine}, vol.~23, no.~12, pp. 16--19, 2008.

\bibitem{ebadi2020lamp}
K.~Ebadi, Y.~Chang, M.~Palieri, A.~Stephens, A.~Hatteland, E.~Heiden,
  A.~Thakur, N.~Funabiki, B.~Morrell, S.~Wood, \emph{et~al.}, ``{LAMP:
  Large-scale autonomous mapping and positioning for exploration of
  perceptually-degraded subterranean environments},'' in \emph{2020 IEEE
  International Conference on Robotics and Automation (ICRA)}, 2020, pp.
  80--86.

\bibitem{ma2019exploiting}
W.-C. Ma, I.~Tartavull, I.~A. B{\^a}rsan, S.~Wang, M.~Bai, G.~Mattyus,
  N.~Homayounfar, S.~K. Lakshmikanth, A.~Pokrovsky, and R.~Urtasun,
  ``{Exploiting Sparse Semantic HD Maps for Self-Driving Vehicle
  Localization},'' in \emph{2019 IEEE/RSJ International Conference on
  Intelligent Robots and Systems (IROS)}, 2019, pp. 5304--5311.

\bibitem{mur2015orb}
R.~Mur-Artal, J.~M.~M. Montiel, and J.~D. Tardos, ``{ORB-SLAM: a versatile and
  accurate monocular SLAM system},'' \emph{IEEE transactions on robotics},
  vol.~31, no.~5, pp. 1147--1163, 2015.

\bibitem{forster2014svo}
C.~Forster, M.~Pizzoli, and D.~Scaramuzza, ``{SVO: Fast semi-direct monocular
  visual odometry},'' in \emph{2014 IEEE international conference on robotics
  and automation (ICRA)}, 2014, pp. 15--22.

\bibitem{qin2018vins}
T.~Qin, P.~Li, and S.~Shen, ``{VINS-Mono: A robust and versatile monocular
  visual-inertial state estimator},'' \emph{IEEE Transactions on Robotics},
  vol.~34, no.~4, pp. 1004--1020, 2018.

\bibitem{izadi2011kinectfusion}
S.~Izadi, D.~Kim, O.~Hilliges, D.~Molyneaux, R.~Newcombe, P.~Kohli, J.~Shotton,
  S.~Hodges, D.~Freeman, A.~Davison, \emph{et~al.}, ``{KinectFusion: real-time
  3D reconstruction and interaction using a moving depth camera},'' in
  \emph{Proceedings of the 24th annual ACM symposium on User interface software
  and technology}, 2011, pp. 559--568.

\bibitem{whelan2015real}
T.~Whelan, M.~Kaess, H.~Johannsson, M.~Fallon, J.~J. Leonard, and J.~McDonald,
  ``Real-time large-scale dense rgb-d slam with volumetric fusion,'' \emph{The
  International Journal of Robotics Research}, vol.~34, no. 4-5, pp. 598--626,
  2015.

\bibitem{labbe2019rtab}
M.~Labb{\'e} and F.~Michaud, ``{RTAB-Map as an open-source lidar and visual
  simultaneous localization and mapping library for large-scale and long-term
  online operation},'' \emph{Journal of Field Robotics}, vol.~36, no.~2, pp.
  416--446, 2019.

\bibitem{zhang2014loam}
J.~Zhang and S.~Singh, ``Loam: Lidar odometry and mapping in real-time.'' in
  \emph{Robotics: Science and Systems}, vol.~2, no.~9, 2014.

\bibitem{deschaud2018imls}
J.-E. Deschaud, ``Imls-slam: scan-to-model matching based on 3d data,'' in
  \emph{IEEE International Conference on Robotics and Automation (ICRA)}.\hskip
  1em plus 0.5em minus 0.4em\relax IEEE, 2018, pp. 2480--2485.

\bibitem{graeter2018limo}
J.~Graeter, A.~Wilczynski, and M.~Lauer, ``Limo: Lidar-monocular visual
  odometry,'' in \emph{IEEE/RSJ International Conference on Intelligent Robots
  and Systems (IROS)}, 2018, pp. 7872--7879.

\bibitem{neuhaus2018mc2slam}
F.~Neuhaus, T.~Ko{\ss}, R.~Kohnen, and D.~Paulus, ``Mc2slam: Real-time inertial
  lidar odometry using two-scan motion compensation,'' in \emph{German
  Conference on Pattern Recognition}, 2018.

\bibitem{shan2018lego}
T.~Shan and B.~Englot, ``Lego-loam: Lightweight and ground-optimized lidar
  odometry and mapping on variable terrain,'' in \emph{2018 IEEE/RSJ
  International Conference on Intelligent Robots and Systems (IROS)}, 2018, pp.
  4758--4765.

\bibitem{behley2018efficient}
J.~Behley and C.~Stachniss, ``Efficient surfel-based slam using 3d laser range
  data in urban environments.'' in \emph{Robotics: Science and Systems}, 2018.

\bibitem{chen2019suma++}
X.~Chen, A.~Milioto, E.~Palazzolo, P.~Gigu{\`e}re, J.~Behley, and C.~Stachniss,
  ``{Suma++: Efficient lidar-based semantic slam},'' in \emph{2019 IEEE/RSJ
  International Conference on Intelligent Robots and Systems (IROS)}, 2019, pp.
  4530--4537.

\bibitem{kovalenko2019sensor}
D.~Kovalenko, M.~Korobkin, and A.~Minin, ``Sensor aware lidar odometry,'' in
  \emph{2019 European Conference on Mobile Robots (ECMR)}, 2019, pp. 1--6.

\bibitem{li2019net}
Q.~Li, S.~Chen, C.~Wang, X.~Li, C.~Wen, M.~Cheng, and J.~Li, ``{LO-Net: Deep
  Real-time Lidar Odometry},'' in \emph{Proceedings of the IEEE Conference on
  Computer Vision and Pattern Recognition}, 2019, pp. 8473--8482.

\bibitem{ye2019tightly}
H.~Ye, Y.~Chen, and M.~Liu, ``Tightly coupled 3d lidar inertial odometry and
  mapping,'' in \emph{2019 International Conference on Robotics and Automation
  (ICRA)}, 2019, pp. 3144--3150.

\bibitem{zuo2019lic}
X.~Zuo, P.~Geneva, W.~Lee, Y.~Liu, and G.~Huang, ``{LIC-Fusion:
  LiDAR-Inertial-Camera Odometry},'' in \emph{2019 IEEE/RSJ International
  Conference on Intelligent Robots and Systems (IROS)}, 2019, pp. 5848--5854.

\bibitem{lin2020livox}
J.~Lin and F.~Zhang, ``Loam livox: A fast, robust, high-precision lidar
  odometry and mapping package for lidars of small fov,'' in \emph{2020 IEEE
  International Conference on Robotics and Automation (ICRA)}, 2020, pp.
  3126--3131.

\bibitem{chen2020sloamforest}
S.~W. Chen, G.~V. Nardari, E.~S. Lee, C.~Qu, X.~Liu, R.~A.~F. Romero, and
  V.~Kumar, ``{SLOAM: Semantic lidar odometry and mapping for forest
  inventory},'' \emph{IEEE Robotics and Automation Letters}, vol.~5, no.~2, pp.
  612--619, 2020.

\bibitem{aloam}
\BIBentryALTinterwordspacing
T.~Qin and S.~Cao, ``A-loam: Advanced implementation of loam,'' 2019. [Online].
  Available: \url{https://github.com/HKUST-Aerial-Robotics/A-LOAM}
\BIBentrySTDinterwordspacing

\bibitem{liosam2020shan}
T.~Shan, B.~Englot, D.~Meyers, W.~Wang, C.~Ratti, and R.~Daniela, ``{LIO-SAM:
  Tightly-coupled Lidar Inertial Odometry via Smoothing and Mapping},'' in
  \emph{IEEE/RSJ International Conference on Intelligent Robots and Systems
  (IROS)}.\hskip 1em plus 0.5em minus 0.4em\relax IEEE, 2020.

\bibitem{garcia2020fail}
I.~Garc{\'\i}a~Daza, M.~Rentero, C.~Salinas~Maldonado, R.~Izquierdo~Gonzalo,
  and N.~Hern{\'a}ndez~Parra, ``Fail-aware lidar-based odometry for autonomous
  vehicles,'' \emph{Sensors}, vol.~20, no.~15, p. 4097, 2020.

\bibitem{zhou2020s4}
B.~Zhou, Y.~He, K.~Qian, X.~Ma, and X.~Li, ``{S4-SLAM: A real-time 3D LIDAR
  SLAM system for ground/watersurface multi-scene outdoor applications},''
  \emph{Autonomous Robots}, pp. 1--22, 2020.

\bibitem{chen2020psf}
G.~Chen, B.~Wang, X.~Wang, H.~Deng, B.~Wang, and S.~Zhang, ``{PSF-LO:
  Parameterized Semantic Features Based Lidar Odometry},'' \emph{arXiv preprint
  arXiv:2010.13355}, 2020.

\bibitem{zheng2020lodonet}
C.~Zheng, Y.~Lyu, M.~Li, and Z.~Zhang, ``Lodonet: A deep neural network with 2d
  keypoint matching for 3d lidar odometry estimation,'' in \emph{Proceedings of
  the 28th ACM International Conference on Multimedia}, 2020, pp. 2391--2399.

\bibitem{he2016m2dp}
L.~He, X.~Wang, and H.~Zhang, ``{M2DP: A novel 3D point cloud descriptor and
  its application in loop closure detection},'' in \emph{2016 IEEE/RSJ
  International Conference on Intelligent Robots and Systems (IROS)}, 2016, pp.
  231--237.

\bibitem{dube2020segmap}
R.~Dub{\'e}, A.~Cramariuc, D.~Dugas, H.~Sommer, M.~Dymczyk, J.~Nieto,
  R.~Siegwart, and C.~Cadena, ``{SegMap: Segment-based mapping and localization
  using data-driven descriptors},'' \emph{The International Journal of Robotics
  Research}, vol.~39, no. 2-3, pp. 339--355, 2020.

\bibitem{kim2018scan}
G.~Kim and A.~Kim, ``Scan context: Egocentric spatial descriptor for place
  recognition within 3d point cloud map,'' in \emph{2018 IEEE/RSJ International
  Conference on Intelligent Robots and Systems (IROS)}, 2018, pp. 4802--4809.

\bibitem{wang2020isc}
H.~{Wang}, C.~{Wang}, and L.~{Xie}, ``Intensity scan context: Coding intensity
  and geometry relations for loop closure detection,'' in \emph{2020 IEEE
  International Conference on Robotics and Automation (ICRA)}, 2020, pp.
  2095--2101.

\bibitem{jiang2020lipmatch}
J.~Jiang, J.~Wang, P.~Wang, P.~Bao, and Z.~Chen, ``{LiPMatch: LiDAR Point Cloud
  Plane Based Loop-Closure},'' \emph{IEEE Robotics and Automation Letters},
  vol.~5, no.~4, pp. 6861--6868, 2020.

\bibitem{liang2020novel}
F.~Liang, B.~Yang, Z.~Dong, R.~Huang, Y.~Zang, and Y.~Pan, ``A novel skyline
  context descriptor for rapid localization of terrestrial laser scans to
  airborne laser scanning point clouds,'' \emph{ISPRS Journal of Photogrammetry
  and Remote Sensing}, vol. 165, pp. 120--132, 2020.

\bibitem{chen2020overlapnet}
X.~Chen, T.~L{\"a}be, A.~Milioto, T.~R{\"o}hling, O.~Vysotska, A.~Haag,
  J.~Behley, C.~Stachniss, and F.~Fraunhofer, ``{OverlapNet: Loop closing for
  LiDAR-based SLAM},'' in \emph{Proc. Robot.: Sci. Syst.}, 2020.

\bibitem{zaganidis2019semantically}
A.~Zaganidis, A.~Zerntev, T.~Duckett, and G.~Cielniak, ``{Semantically Assisted
  Loop Closure in SLAM Using NDT Histograms},'' in \emph{2019 IEEE/RSJ
  International Conference on Intelligent Robots and Systems (IROS)}, 2019, pp.
  4562--4568.

\bibitem{kaess2008isam}
``{iSAM: Incremental smoothing and mapping}, author={Kaess, Michael and
  Ranganathan, Ananth and Dellaert, Frank},'' \emph{IEEE Transactions on
  Robotics}, vol.~24, no.~6, pp. 1365--1378, 2008.

\bibitem{grisetti2010hierarchical}
G.~Grisetti, R.~K{\"u}mmerle, C.~Stachniss, U.~Frese, and C.~Hertzberg,
  ``{Hierarchical optimization on manifolds for online 2D and 3D mapping},'' in
  \emph{2010 IEEE International Conference on Robotics and Automation}, 2010,
  pp. 273--278.

\bibitem{ni2010multilevelsubmap}
K.~Ni and F.~Dellaert, ``Multi-level submap based slam using nested
  dissection,'' in \emph{2010 IEEE/RSJ International Conference on Intelligent
  Robots and Systems}, 2010, pp. 2558--2565.

\bibitem{kummerle2011g}
R.~K{\"u}mmerle, G.~Grisetti, H.~Strasdat, K.~Konolige, and W.~Burgard, ``g2o:
  A general framework for graph optimization,'' in \emph{2011 IEEE
  International Conference on Robotics and Automation}, 2011, pp. 3607--3613.

\bibitem{grisetti2012robustoptimizationofgraph}
G.~Grisetti, R.~K{\"u}mmerle, and K.~Ni, ``Robust optimization of factor graphs
  by using condensed measurements,'' in \emph{2012 IEEE/RSJ International
  Conference on Intelligent Robots and Systems}, 2012, pp. 581--588.

\bibitem{Blanco-Claraco-RSS-19}
J.~L. Blanco-Claraco, ``A modular optimization framework for localization and
  mapping,'' in \emph{Proceedings of Robotics: Science and Systems},
  FreiburgimBreisgau, Germany, June 2019.

\bibitem{li2020urbanslamisprs}
S.~Li, G.~Li, L.~Wang, and Y.~Qin, ``Slam integrated mobile mapping system in
  complex urban environments,'' \emph{ISPRS Journal of Photogrammetry and
  Remote Sensing}, vol. 166, pp. 316--332, 2020.

\bibitem{zhao2019highway}
S.~Zhao, Z.~Fang, H.~Li, and S.~Scherer, ``A robust laser-inertial odometry and
  mapping method for large-scale highway environments,'' in \emph{2019 IEEE/RSJ
  International Conference on Intelligent Robots and Systems (IROS)}, 2019, pp.
  1285--1292.

\bibitem{palieri2020locus}
M.~Palieri, B.~Morrell, A.~Thakur, K.~Ebadi, J.~Nash, A.~Chatterjee,
  C.~Kanellakis, L.~Carlone, C.~Guaragnella, and A.-a. Agha-mohammadi,
  ``{LOCUS: A Multi-Sensor Lidar-Centric Solution for High-Precision Odometry
  and 3D Mapping in Real-Time},'' \emph{IEEE Robotics and Automation Letters},
  vol.~6, no.~2, pp. 421--428, 2020.

\bibitem{Yang20tro-teaser}
H.~Yang, J.~Shi, and L.~Carlone, ``{TEASER: Fast and Certifiable Point Cloud
  Registration},'' \emph{{IEEE} Trans. Robotics}, 2020.

\bibitem{besl1992method}
P.~J. Besl and N.~D. McKay, ``Method for registration of 3-d shapes,'' in
  \emph{Sensor fusion IV}, vol. 1611, 1992, pp. 586--606.

\bibitem{rusinkiewicz2001pointplaneicp}
S.~Rusinkiewicz and M.~Levoy, ``Efficient variants of the icp algorithm,'' in
  \emph{Proceedings third international conference on 3-D digital imaging and
  modeling}, 2001, pp. 145--152.

\bibitem{segal2009generalized}
A.~Segal, D.~Haehnel, and S.~Thrun, ``{Generalized-ICP},'' in \emph{Robotics:
  science and systems}, vol.~2, no.~4, 2009, p. 435.

\bibitem{censi2008icp}
A.~Censi, ``An icp variant using a point-to-line metric,'' in \emph{IEEE
  International Conference on Robotics and Automation}, 2008, pp. 19--25.

\bibitem{yokozuka2021litamin2}
M.~Yokozuka, K.~Koide, S.~Oishi, and A.~Banno, ``{LiTAMIN2: Ultra Light
  LiDAR-based SLAM using Geometric Approximation applied with KL-Divergence},''
  \emph{IEEE International Conference on Robotics and Automation (ICRA)}, 2021.

\bibitem{low2004linear}
K.-L. Low, ``Linear least-squares optimization for point-to-plane icp surface
  registration,'' \emph{Chapel Hill, University of North Carolina}, vol.~4,
  no.~10, pp. 1--3, 2004.

\bibitem{pomerleau2015review}
F.~Pomerleau, F.~Colas, and R.~Siegwart, ``A review of point cloud registration
  algorithms for mobile robotics,'' \emph{Foundations and Trends in Robotics},
  vol.~4, no.~1, pp. 1--104, 2015.

\bibitem{kuhner2020large}
T.~K{\"u}hner and J.~K{\"u}mmerle, ``Large-scale volumetric scene
  reconstruction using lidar,'' in \emph{IEEE International Conference on
  Robotics and Automation (ICRA)}, 2020, pp. 6261--6267.

\bibitem{rusu2009fpfh}
R.~B. Rusu, N.~Blodow, and M.~Beetz, ``{Fast point feature histograms (FPFH)
  for 3D registration},'' in \emph{2009 IEEE international conference on
  robotics and automation}, 2009, pp. 3212--3217.

\bibitem{huang2020predator}
S.~Huang, Z.~Gojcic, M.~Usvyatsov, A.~Wieser, and K.~Schindler, ``{PREDATOR:
  Registration of 3D Point Clouds with Low Overlap},'' in \emph{Conference on
  Computer Vision and Pattern Recognition (CVPR)}, 2021.

\bibitem{yang2020graduated}
H.~Yang, P.~Antonante, V.~Tzoumas, and L.~Carlone, ``Graduated non-convexity
  for robust spatial perception: From non-minimal solvers to global outlier
  rejection,'' \emph{IEEE Robotics and Automation Letters}, vol.~5, no.~2, pp.
  1127--1134, 2020.

\bibitem{mellado2014super}
N.~Mellado, D.~Aiger, and N.~J. Mitra, ``{Super 4pcs fast global pointcloud
  registration via smart indexing},'' in \emph{Computer Graphics Forum},
  vol.~33, no.~5, 2014, pp. 205--215.

\bibitem{yang2015go}
J.~Yang, H.~Li, D.~Campbell, and Y.~Jia, ``{Go-ICP: A globally optimal solution
  to 3D ICP point-set registration},'' \emph{IEEE transactions on pattern
  analysis and machine intelligence}, vol.~38, no.~11, pp. 2241--2254, 2015.

\bibitem{cai2019practical}
Z.~Cai, T.-J. Chin, A.~P. Bustos, and K.~Schindler, ``{Practical optimal
  registration of terrestrial LiDAR scan pairs},'' \emph{ISPRS journal of
  photogrammetry and remote sensing}, vol. 147, pp. 118--131, 2019.

\bibitem{hackel2016fast}
T.~Hackel, J.~D. Wegner, and K.~Schindler, ``{Fast semantic segmentation of 3D
  point clouds with strongly varying density},'' \emph{ISPRS annals of the
  photogrammetry, remote sensing and spatial information sciences}, vol.~3, pp.
  177--184, 2016.

\bibitem{barron2019general}
J.~T. Barron, ``A general and adaptive robust loss function,'' in
  \emph{Proceedings of the IEEE Conference on Computer Vision and Pattern
  Recognition}, 2019, pp. 4331--4339.

\bibitem{chebrolu2020adaptive}
N.~Chebrolu, T.~L{\"a}be, O.~Vysotska, J.~Behley, and C.~Stachniss, ``Adaptive
  robust kernels for non-linear least squares problems,'' \emph{arXiv preprint
  arXiv:2004.14938}, 2020.

\bibitem{geiger2012we}
A.~Geiger, P.~Lenz, and R.~Urtasun, ``Are we ready for autonomous driving? the
  kitti vision benchmark suite,'' in \emph{IEEE Conference on Computer Vision
  and Pattern Recognition (CVPR)}, 2012.

\bibitem{wang2020isprs}
C.~Wang, Y.~Dai, N.~Elsheimy, C.~Wen, G.~Retscher, Z.~Kang, and A.~Lingua,
  ``Isprs benchmark on multisensory indoor mapping and positioning,''
  \emph{ISPRS Annals of the Photogrammetry, Remote Sensing and Spatial
  Information Sciences}, vol.~5, pp. 117--123, 2020.

\bibitem{yin2020caelo}
D.~Yin, Q.~Zhang, J.~Liu, X.~Liang, Y.~Wang, J.~Maanpää, H.~Ma, J.~Hyyppä,
  and R.~Chen, ``{CAE-LO: LiDAR Odometry Leveraging Fully Unsupervised
  Convolutional Auto-Encoder for Interest Point Detection and Feature
  Description},'' \emph{arXiv preprint arXiv:2001.01354}, 2020.

\end{thebibliography}
	
\end{document}